\documentclass{article}

\usepackage{microtype}
\usepackage{graphicx}
\usepackage{dblfloatfix}
\usepackage{booktabs} 

\usepackage{hyperref}
\usepackage{url}

\usepackage{booktabs}       
\usepackage{amsfonts}       
\usepackage{amsmath,amsthm}
\usepackage{multirow}
\usepackage{wrapfig}
\usepackage{float}

\usepackage{nicefrac}       
\usepackage{microtype}      
\usepackage{color}

\usepackage{xcolor}     
\usepackage{pgfplots}
\usepackage{subcaption}
\usepackage{sidecap}
\usepackage{cleveref}

\newtheorem{Def}{Definition}
\newtheorem{Lem}{Lemma}
\newtheorem{proposition}{Proposition}
\newcommand{\methodName}{\textit{Bayesian Offline Reinforcement Learning (BOReL)}}
\newcommand{\acronym}{BOReL}

\usepackage[accepted]{icml2021}

\icmltitlerunning{Offline Meta Learning of Exploration}

\begin{document}

\twocolumn[
\icmltitle{Offline Meta Learning of Exploration}



\icmlsetsymbol{equal}{*}

\begin{icmlauthorlist}
\icmlauthor{Ron Dorfman}{Technion}
\icmlauthor{Idan Shenfeld}{Technion}
\icmlauthor{Aviv Tamar}{Technion}
\end{icmlauthorlist}

\icmlaffiliation{Technion}{Department of Electrical Engineering, Technion, Israel}

\icmlcorrespondingauthor{Ron Dorfman}{rdorfman@campus.technion.ac.il}

\icmlkeywords{Machine Learning, ICML}

\vskip 0.3in
]



\printAffiliationsAndNotice{}  

\begin{abstract}
Consider the following instance of the Offline Meta Reinforcement Learning (OMRL) problem: given the complete training logs of $N$ conventional RL agents, trained on $N$ different tasks, design a meta-agent that can quickly maximize reward in a new, unseen task from the same task distribution. In particular, while each conventional RL agent explored and exploited its own different task, the meta-agent must identify regularities in the data that lead to effective exploration/exploitation in the unseen task. Here, we take a Bayesian RL (BRL) view, and seek to learn a Bayes-optimal policy from the offline data. Building on the recent VariBAD BRL approach, we develop an off-policy BRL method that learns to plan an  exploration strategy based on an adaptive neural belief estimate. However, learning to infer such a belief from offline data brings a new identifiability issue we term MDP ambiguity. We characterize the problem, and suggest resolutions via data collection and modification procedures.
Finally, we evaluate our framework on a diverse set of domains, including difficult sparse reward tasks, and demonstrate learning of effective exploration behavior that is qualitatively different from the exploration used by any RL agent in the data. 
\end{abstract}

\section{Introduction}
A central question in reinforcement learning (RL) is how to learn quickly (i.e., with few samples) in a new environment. Meta-RL addresses this issue by assuming a distribution over possible environments, and having access to a large set of environments from this distribution during training~\cite{duan2016rl, finn2017model}. Intuitively, the meta-RL agent can learn regularities in the environments, which allow quick learning in any environment that shares a similar structure. Indeed, recent work demonstrated this by training memory-based controllers that `identify' the domain~\citep{duan2016rl, rakelly2019efficient, humplik2019meta}, or by learning a parameter initialization that leads to good performance within a few gradient steps~\citep{finn2017model}.

Another formulation of quick RL is Bayesian RL~\citep[BRL,][]{ghavamzadeh2016bayesian}. In BRL, the environment parameters are treated as unobserved variables, with a known prior distribution. Consequentially, the standard problem of maximizing expected returns (taken with respect to the posterior distribution) \emph{explicitly accounts for the environment uncertainty}, and its solution is a \emph{Bayes-optimal} policy, wherein actions optimally balance exploration and exploitation. Recently, \citet{zintgraf2020varibad} showed that meta-RL is in fact an instance of BRL, where the meta-RL environment distribution is simply the BRL prior. Furthermore, a Bayes-optimal policy can be trained using standard policy gradient methods, simply by adding to the state the posterior belief over the environment parameters. The VariBAD algorithm~\citep{zintgraf2020varibad} is an implementation of this approach that uses a variational autoencoder (VAE) for adaptive belief estimation and deep neural network policies.

Most meta-RL studies, including VariBAD, have focused on the \emph{online} setting, where, during training, the meta-RL policy is continually updated using data collected from running it in the training environments. In domains where data collection is expensive, such as robotics and healthcare to name a few, online training is a limiting factor. For standard RL, offline (a.k.a. batch) RL mitigates this problem by learning from data collected beforehand by an arbitrary policy~\citep{ernst2005tree,levine2020offline}. In this work we investigate the \emph{offline approach to meta-RL} (OMRL).

\begin{figure*}[t]
        \vspace{-1.5em}
        \begin{subfigure}{0.5\textwidth}
        \centering
          \includegraphics[draft=false,width=0.75\linewidth]{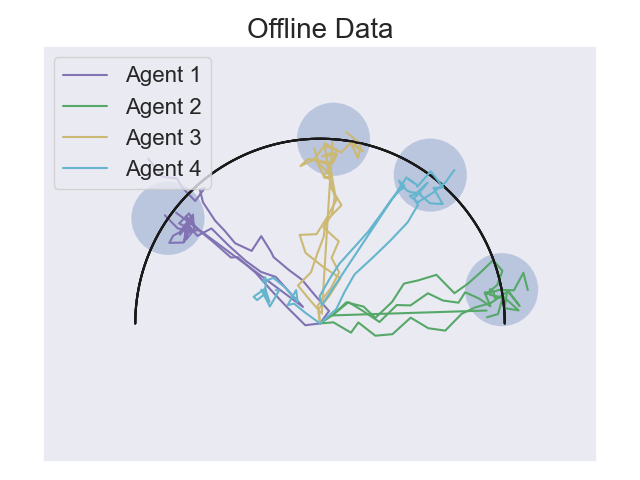} 
        \label{fig:subim1} 
        \end{subfigure}\hfill
        \begin{subfigure}{0.5\textwidth}
            \centering
            \includegraphics[draft=false,width=0.75\linewidth]{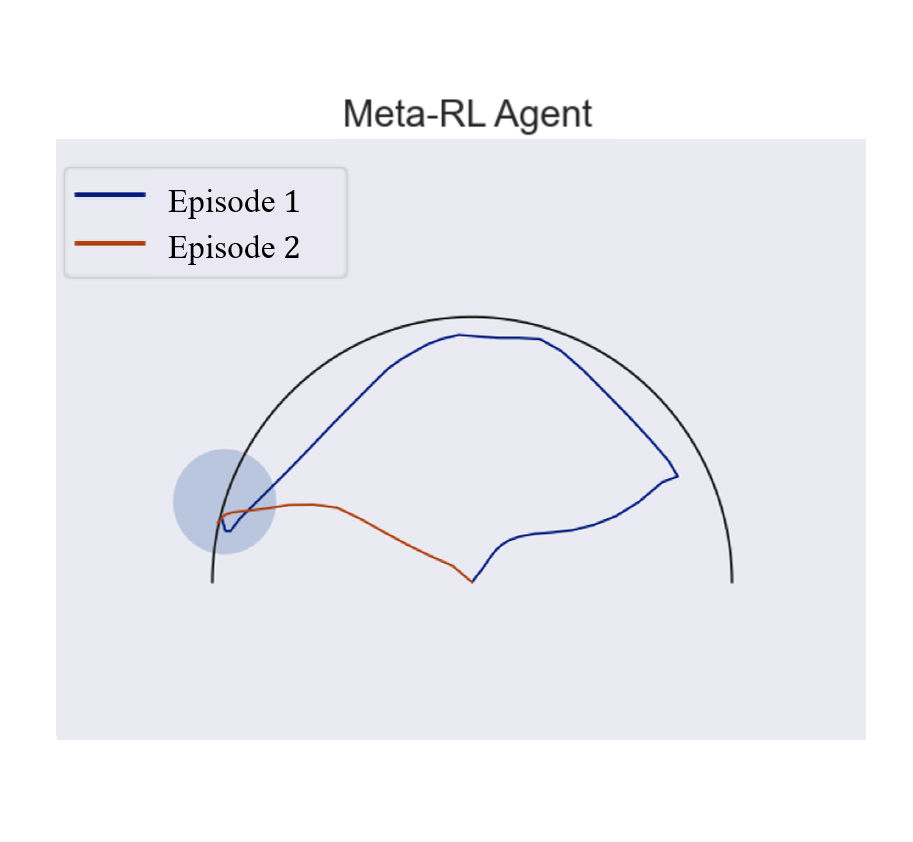}
            \label{fig:subim2}
        \end{subfigure}
    \vspace{-2.5em}
    \caption{Offline meta-RL on the Semi-Circle domain: the task is to navigate to a goal position that can be anywhere on the semi-circle. The reward is sparse (light-blue), and the offline data (left) contains training logs of conventional RL agents trained to find individual goals. The meta-RL agent (right) needs to find a policy that quickly finds the unknown goal, here, by searching across the semi-circle in the first episode, and directly reaching it the second -- a completely different strategy from the dominant behaviors in the data.}
    \label{fig:illustration}
    \vspace{-1.0em}
\end{figure*}

Any offline RL approach is heavily influenced by the data collection policy. To ground our investigation, we focus on the following practical setting: we assume that data has been collected by running standard RL agents on a set of environments from the environment distribution. 
While the data was not specifically collected for the meta-RL task, we hypothesize that regularities between the training domains can still be learned, to provide faster learning in new environments. Figure \ref{fig:illustration} illustrates our problem: in this navigation task, each RL agent in the data learned to find its own goal, and converged to a behavior that quickly navigates toward it. The meta-RL agent, on the other hand, needs to learn a completely different behavior that effectively \textit{searches} for the unknown goal position.

Our method for solving OMRL is an off-policy variant of the VariBAD algorithm, based on replacing the on-policy policy gradient optimization in VariBAD with an off-policy Q-learning based approach. This, however, requires some care, as Q-learning applies to states of fully observed systems. We show that the VariBAD approach of augmenting states with the belief in the data applies to the off-policy setting as well, leading to an effective and practical algorithm. The offline setting, however, brings about another challenge -- when the agent visits different parts of the state space in different environments, learning to identify the correct environment and obtain an accurate belief estimate becomes challenging, a problem we term \textit{MDP ambiguity}. We formalize this problem, and discuss how it manifests in common scenarios such as sparse rewards or sparse differences in transitions. Based on our formalization, we propose a general data collection strategy that can mitigate the problem. Further, when ambiguity is only due to reward differences, we show that a simple reward relabelling trick suffices, without changing data collection. We collectively term our data collection/relabelling and off-policy algorithm as \methodName. 

In our experiments, we show that \acronym\ learns effective exploration policies from offline data on both discrete and continuous control problems. We demonstrate significantly better exploration than meta-RL methods based on Thompson sampling such as PEARL~\citep{rakelly2019efficient}, \textit{even when these methods are allowed to train online}. Furthermore, we explore the issue of MDP ambiguity in practice, and demonstrate that, when applicable, our proposed solutions successfully mitigate it.

Our main contributions are as follows: to our knowledge, this is the first study of meta learning exploration in the offline setting; we provide the necessary theory to extend VariBAD to off-policy RL; we formulate MDP ambiguity, which characterizes which problems are solvable under the offline BRL setting, and based on this formulation we propose several principled data collection strategies; 
we show non-trivial empirical results that demonstrate significantly better exploration than meta-RL methods based on Thompson sampling; finally, and of independent interest, our off-policy algorithm significantly improves the sample efficiency of conventional VariBAD in the online setting.


\vspace{-0.5em}
\section{Background}
\vspace{-0.5em}
We recapitulate meta-RL, BRL and the VariBAD algorithm.

\textbf{Meta-RL:} In meta-RL, a distribution over tasks is assumed. A task $\mathcal{T}_i$ is described by a Markov Decision Process \citep[MDP,][]{bertsekas1995dynamic} $\mathcal{M}_i = (\mathcal{S}, \mathcal{A}, \mathcal{R}_i, \mathcal{P}_i)$, where the state space $\mathcal{S}$ and the action space $\mathcal{A}$ are shared across tasks, and $\mathcal{R}_i$ and $\mathcal{P}_i$ are task specific reward and transition functions. Thus, we write the task distribution as $p(\mathcal{R},\mathcal{P})$. For simplicity, we assume throughout that the initial state distribution $P_{init}(s_0)$ is the same for all MDPs. The goal in meta-RL is to train an agent that can quickly maximize reward in new, unseen tasks, drawn from  $p(\mathcal{R},\mathcal{P})$. 

\textbf{Bayesian Reinforcement Learning:}
The goal in BRL is to find the optimal policy $\pi$ in an MDP, when the transitions and rewards are not known in advance. Similar to meta-RL, we assume a prior over the MDP parameters $p(\mathcal{R},\mathcal{P})$, and seek to maximize 
the expected discounted return,
\vspace{-0.5em}
\begin{equation}\label{eq:BRL_objective}
    \mathbb{E}_\pi \left[\sum_{t=0}^\infty \gamma^t r(s_t,a_t)\right], 
    \vspace{-0.5em}
\end{equation}
where the expectation is taken with respect to \emph{both the uncertainty in state-action transitions} $s_{t+1}\sim \mathcal{P}(\cdot|s_t,a_t)$, $a_t\sim \pi,$ \emph{and the uncertainty in the MDP parameters} $\mathcal{R},\mathcal{P}\sim p(\mathcal{R},\mathcal{P})$.\footnote{For ease of presentation, we consider the infinite horizon discounted return. Our formulation easily extends to the episodic and finite horizon settings, as considered in our experiments.} 
Key here is that this formulation naturally accounts for the exploration/exploitation tradeoff -- an optimal agent must plan its actions to reduce uncertainty in the MDP parameters, if such leads to higher rewards.

One way to approach the BRL problem is to model $\mathcal{R},\mathcal{P}$ as unobserved state variables in a partially observed MDP \citep[POMDP,][]{cassandra1994acting}, reducing the problem to solving a particular POMDP instance where the unobserved variables do not change in time. 
The \emph{belief} at time $t$, $b_t$, denotes the posterior probability over $\mathcal{R},\mathcal{P}$ given the history of state transitions and rewards observed until this time $b_t=P(\mathcal{R},\mathcal{P}|h_{:t})$, where $h_{:t}=\left\{s_0,a_0,r_1,s_1\dots,r_t,s_t \right\}$ (note that we denote the reward after observing the state and action at time $t$ as $r_{t+1}=r(s_t,a_t)$). The belief can be updated iteratively according to Bayes rule, where $b_0(\mathcal{R},\mathcal{P})=p(\mathcal{R},\mathcal{P})$, and:
$
    b_{t+1}(\mathcal{R},\mathcal{P}) = P(\mathcal{R},\mathcal{P}|h_{:t+1})\propto P(s_{t+1},r_{t+1}|h_{:t},\mathcal{R},\mathcal{P}) b_t(\mathcal{R},\mathcal{P}).
$

Similar to the idea of solving a POMDP by representing it as an MDP over belief states, the state in BRL can be augmented with the belief to result in the Bayes-Adaptive MDP model \citep[BAMDP,][]{duff2002optimal}. Denote the augmented state $s_t^+ = (s_t, b_t)$ and the augmented state space $\mathcal{S}^+ = \mathcal{S} \times \mathcal{B}$, where $\mathcal{B}$ denotes the belief space. The transitions in the BAMDP are given by:
$
    P^+(s_{t+1}^+|s_{t}^+,a_t) = \mathbb{E}_{b_t}\left[ \mathcal{P}(s_{t+1}|s_t,a_t)\right] \delta\left( b_{t+1} = P(\mathcal{R},\mathcal{P}|h_{:t+1})\right),
$
and the reward in the BAMDP is the expected reward with respect to the belief:
$
    R^+(s_{t}^+, a_t) = \mathbb{E}_{b_{t}}\left[ \mathcal{R}(s_{t},a_t)\right] .
$
The Bayes-optimal agent seeks to maximize the expected discounted return in the BAMDP, and the optimal solution of the BAMDP gives the optimal BRL policy.
As in standard MDPs, the optimal action-value function in the BAMDP satisfies the Bellman equation: $\forall s^+ \in \mathcal{S^+}, a\in \mathcal{A}$ we have that
\vspace{-0.5em}
\begin{equation}\label{eq:Bellman_Q}
    Q(s^+\!,\!a) \!= \!R^+(s^+,a) + \gamma\mathbb{E}_{s^{+'}\sim P^{+}}\big[\max_{a'}{Q(s^{+'}\!,a')}\big].
    \vspace{-0.5em}
\end{equation}
Computing a Bayes-optimal agent amounts to solving the BAMDP, where the optimal policy is a function of the augmented state. For most problems this is intractable, as the augmented state space is continuous and high-dimensional, and the posterior update is also intractable in general. 

\textbf{The VariBAD Algorithm:}
VariBAD~\citep{zintgraf2020varibad} approximates the Bayes-optimal solution by combining a model for the MDP parameter uncertainty, and an optimization method for the corresponding BAMDP. The MDP parameters are represented by a vector $m\in\mathbb{R}^d$, corresponding to the latent variables in a parametric generative model for the state-reward trajectory distribution conditioned on the actions $P(s_0,r_1,s_1\ldots,r_H, s_H|a_0,\ldots,a_{H-1}) = \int p_\theta(m)p_\theta(s_0,r_1,s_1\ldots,r_H, s_H|m, a_0,\ldots,a_{H-1})dm$. The model parameters $\theta$ are learned by a variational approximation to the maximum likelihood objective, where the variational approximation to the posterior $P(m|s_0,r_1,s_1\ldots,r_H, s_H, a_0,\ldots,a_{H-1})$ is chosen to have the structure $q_\phi(m|s_0,a_0,r_1,s_1\ldots,r_t, s_t) = q_{\phi}(m|h_{:t})$. That is, the approximate posterior is conditioned on the history up to time $t$. The evidence lower bound (ELBO) in this case is 
$
    ELBO_t = \mathbb{E}_{m\sim q_{\phi}(\cdot|h_{:t})}\left[ \log{p_\theta(s_0,r_1,s_1\ldots,r_H, s_H|m, a_0,\ldots,a_{H-1})}\right] - D_{KL}(q_{\phi}(m|h_{:t}) || p_{\theta}(m)).
$
The main claim of \citet{zintgraf2020varibad} is that $q_\phi(m|h_{:t})$ can be taken as an approximation of the belief $b_t$. 
In practice, $q_{\phi}(m|h_{:t})$ is represented as a Gaussian distribution $q(m|h_{:t})=\mathcal{N}(\mu(h_{:t}),\Sigma(h_{:t}))$, where $\mu$ and $\Sigma$ are learned recurrent neural networks. While other neural belief representations could be used~\citep[][]{guo2018neural}, we chose to focus on VariBAD for concreteness.

To approximately solve the BAMDP, \citet{zintgraf2020varibad} exploit the fact that an optimal BAMDP policy is a function of the state and belief, and therefore consider neural network policies that take the augmented BAMDP state as input $\pi(a_t | s_t,q_{\phi}(m|h_{:t}))$, where the posterior is practically represented by the distribution parameters $\mu(h_{:t}),\Sigma(h_{:t})$. The policies are trained using policy gradients, optimizing
\begin{equation}\label{eq:BRL_PG_Objective}
J(\pi) = \mathbb{E_{\mathcal{R},\mathcal{P}}} \mathbb{E_{\pi}} \left[\sum_{t=0}^H \gamma^t r(s_t,a_t)\right].
\end{equation}
The expectation over MDP parameters in (\ref{eq:BRL_PG_Objective}) is approximated by averaging over training environments, and the RL agent is trained online, alongside the VAE. 

\vspace{-0.5em}
\section{OMRL and Off-Policy VariBAD}
In this section, we derive an off-policy variant of the VariBAD algorithm, and apply it to the OMRL problem. We first describe OMRL, and then present our algorithm.

\vspace{-0.5em}
\subsection{OMRL Problem Definition}\label{ssec:omrl_problem_def}
We follow the Meta-RL and BRL formulation described above, with a prior distribution over MDP parameters $p(\mathcal{R},\mathcal{P})$. We are provided training data of an agent interacting with $N$ different MDPs, $\left\{\mathcal{R}_i,\mathcal{P}_i\right\}_{i=1}^N,$ sampled from the prior. We assume that each interaction is organized as $M$ trajectories of length $H$, $\tau^{i,j} = s_0^{i,j}, a_0^{i,j}, r_1^{i,j},s_1^{i,j}\dots, r_H^{i,j}, s_H^{i,j}, \quad i\in 1,\dots,N, j\in 1,\dots,M$, where the rewards satisfy $r_{t+1}^{i,j} = \mathcal{R}_i(s_t^{i,j},a_t^{i,j})$, the transitions satisfy $s_{t+1}^{i,j} \sim \mathcal{P}_i(\cdot | s_t^{i,j},a_t^{i,j})$, and the actions are chosen from an arbitrary data collection policy. To ground our work in a specific context, we sometimes assume that the trajectories are obtained from running a conventional RL agent in each one of the MDPs (i.e., the complete RL training logs), which implicitly specifies the data collection policy. We will later investigate implications of this assumption, 
but emphasize that this is merely an illustration, and our approach does not place any such constraint -- the trajectories can also be collected differently. 
Our goal is to use the data for learning a Bayes-optimal policy, i.e., a policy $\pi$ that maximizes Eq.~(\ref{eq:BRL_objective}).

\subsection{Off-Policy VariBAD} \label{sec:off_policy_varibad}
The online VariBAD algorithm updates the policy using \emph{trajectories sampled from the current policy}, and thus cannot be applied to our offline setting. Our first step is to modify VariBAD to work off-policy.
We start with an observation about the use of the BAMDP formulation in VariBAD, which will motivate our subsequent development. 
\vspace{-1em}
\paragraph{Does VariBAD really optimize the BAMDP?} 
Recall that a BAMDP is in fact a reduction of a POMDP to an MDP over augmented states $s^+ = (s, b)$, and with the rewards and transitions given by $R^+$ and $P^+$. Thus, an optimal Markov policy for the BAMDP exists in the form of $\pi(s^+)$. The VariBAD policy, as described above, similarly takes as input the augmented state, and is thus capable of representing an optimal BAMDP policy. However, \emph{VariBAD's policy optimization in Eq. (\ref{eq:BRL_PG_Objective}) does not make use of the BAMDP parameters} $R^+$ and $P^+$! While at first this may seem counterintuitive, Eq. (\ref{eq:BRL_PG_Objective}) is in fact a sound objective for the BAMDP, as we now show\footnote{This result is closely related to the discussion in~\cite{ortega2019meta}, here applied to our particular setting.}.

\begin{proposition}\label{prop:bamdp}
Let $\tau=s_0,a_0,r_1,s_1\dots, r_H, s_H$ denote a random trajectory from a fixed history dependent policy $\pi$, generated according to the following process. First, MDP parameters $\mathcal{R},\mathcal{P}$ are drawn from the prior $p(\mathcal{R},\mathcal{P})$. Then, the state trajectory is generated according to  $s_0\sim P_{init}$, $a_t\sim \pi(\cdot | s_0,a_0,r_1,\dots,s_t)$, $s_{t+1}\sim \mathcal{P}(\cdot|s_t,a_t)$ and $r_{t+1}\sim \mathcal{R}(s_t,a_t)$. Let $b_t$ denote the posterior belief at time t, $b_t = P(\mathcal{R},\mathcal{P} | s_0,a_0,r_1,\dots,s_t)$. Then 
\vspace{-0.5em}
\begin{equation*}
    \begin{split}
        &P(s_{t+1}|s_0,a_0,r_1,\dots,r_t, s_t, a_t)=\mathbb{E}_{\mathcal{R},\mathcal{P}\sim b_t} \mathcal{P}(s_{t+1}|s_t, a_t),\\
        &\text{and, }P(r_{t+1}\!|\!s_0,a_0,r_1,\!\dots\!,s_t, a_t)\!=\!\mathbb{E}_{\mathcal{R},\mathcal{P}\sim b_t} \mathcal{R}(r_{t+1}|s_t, a_t).
    \end{split}
\end{equation*}
\end{proposition}
\vspace{-0.5em}
For on-policy VariBAD, Proposition \ref{prop:bamdp} shows that the rewards and transitions in each trajectory can be seen as sampled from a distribution that \textbf{in expectation} is equal to $R^+$ and $P^+$, and therefore maximizing Eq.~\ref{eq:BRL_PG_Objective} is valid.\footnote{To further clarify, if we could calculate $R^{+}$, replacing all rewards in the trajectories with $R^{+}$ will result in a lower variance policy update, similar to expected SARSA \citep{van2009theoretical}. }   However, off-policy RL does not take as input trajectories, but tuples of the form $(s,a,r,s')\equiv(state, action, reward, next \ state)$, where states and actions can be sampled \textbf{from any distribution}. For an arbitrary distribution of augmented states, we must replace the rewards and transitions in our data with $R^+$ and $P^+$, which can be difficult to compute. Fortunately, Proposition \ref{prop:bamdp} shows that when collecting data by sampling complete trajectories, this is not necessary, as in expectation, the rewards and transitions are correctly sampled from the BAMDP. In the following, we therefore focus on settings where data can be collected that way, for example, by collecting logs of RL agents trained on the different training tasks.



Based on Proposition \ref{prop:bamdp}, we can use a state augmentation method similar to VariBAD, which we refer to as \textbf{state relabelling}. Consider each trajectory in our data $\tau^{i,j} = s_0^{i,j}, a_0^{i,j}, r_1^{i,j},\dots, s_H^{i,j}$, as defined above. Recall that the VariBAD VAE encoder provides an estimate of the belief given the state history $q(m|h_{:t})=\mathcal{N}(\mu(h_{:t}),\Sigma(h_{:t}))$. Thus, we can run the encoder on every partial $t$-length history $\tau^{i,j}_{:t}$ to obtain the belief at each time step. Following the BAMDP formulation, we define the augmented state $s_t^{+,i,j} = ( s_t^{i,j}, b_t^{i,j} )$, where $b_t^{i,j} = \mu(\tau^{i,j}_{:t}),\Sigma(\tau^{i,j}_{:t})$. We next replace each state in our data $s_t^{i,j}$ with $s_t^{+,i,j}$, effectively transforming the data to as coming from a BAMDP. 
After applying state relabelling, any off-policy RL algorithm can be applied to the modified data, for learning a Bayes-optimal policy. In our experiments we used DQN~\citep{mnih2015human} for discrete action domains, and soft actor critic \citep[SAC,][]{haarnoja2018soft} for continuous control.

\section{Identifiability Problems in OMRL}\label{sec:indentifiability}
\begin{figure}
    \centering
    \includegraphics[width=0.4\textwidth]{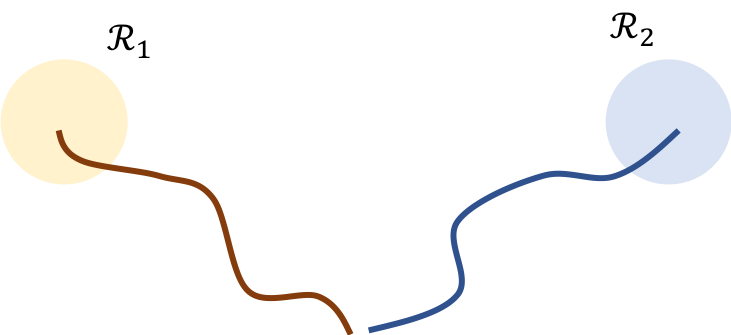}
    \caption{Reward ambiguity: from the two trajectories, it is impossible to know if there are two MDPs with different rewards (blue and yellow circles), or one MDP with rewards at both locations.}
    \label{fig:r_ambiguity}
    \vspace{-2em}
\end{figure}

We take a closer look at the OMRL problem. While in principle, it is possible to simply run off-policy VariBAD on the offline data, we claim that in many problems this may not work well. The reason is that the  VariBAD belief update should reason about the uncertainty in the MDP parameters, which requires to effectively distinguish between the different possible MDPs.
Training the VAE to distinguish between MDPs, however, \textit{depends on the offline data}, and might not always be possible. This problem, which we term \textit{MDP ambiguity}, is illustrated in Figure \ref{fig:r_ambiguity}: consider two MDPs, one with rewards in the blue circle, and the other with rewards in the yellow circle. If the data contains trajectories similar to the ones in the figure, it is impossible to distinguish between having two different MDPs with the indicated rewards, or a single MDP with rewards at both the blue and yellow circles. Accordingly, we cannot expect to learn a meaningful belief update. In the following, we formalize MDP ambiguity, and how it can be avoided.

For an MDP defined by $\{\mathcal{R}, \mathcal{P}\}$, denote by $P_{\mathcal{R}, \mathcal{P}, \pi}(s, a, r, s')$ and $P_{\mathcal{P}, \pi}(s, a)$ the distribution over $(s,a,r,s')$ and $(s,a)$, respectively, induced by a policy $\pi$.

\begin{Def}[MDP Ambiguity]\label{def:ambiguity} Consider data coming from a set of $N$ different MDPs $M = \{\mathcal{R}_i, \mathcal{P}_i\}_{i=1}^N \subset \mathcal{M}$, where $\mathcal{M}$ is an hypothesis set of possible MDPs, and corresponding data collection policies $\{\pi_{\beta}^i\}_{i=1}^N$, resulting in $N$ \textit{different} data distributions $D = \{P_{\mathcal{R}_i, \mathcal{P}_i, \pi_{\beta}^{i}}(s, a, r, s')\}_{i=1}^N$. We say that the data is ambiguous if there is an MDP $\{\mathcal{R},\mathcal{P}\}\in \mathcal{M}$ and two policies $\pi$ and $\pi'$ such that $P_{\mathcal{R}_i, \mathcal{P}_i, \pi_{\beta}^{i}}(s, a, r, s') = P_{\mathcal{R}, \mathcal{P}, \pi}(s, a, r, s')$ and $P_{\mathcal{R}_j, \mathcal{P}_j, \pi_{\beta}^{j}}(s, a, r, s') = P_{\mathcal{R}, \mathcal{P}, \pi'}(s, a, r, s')$, for some $i \neq j$. Otherwise, the data is termed identifiable.\footnote{$P(\cdot)=P'(\cdot)$ means equality almost everywhere; $P(\cdot)\neq P'(\cdot)$ means that equality almost everywhere does not hold.}
\end{Def}

The essence of identifiability, as expressed in Definition \ref{def:ambiguity}, is that there is no single MDP in the hypothesis set that can explain data from multiple MDPs in the data, as in this case it will be impossible to learn an inference model that accurately distinguishes between the different MDPs, even with infinite data.\footnote{For simplicity, Definition \ref{def:ambiguity} considers a discrete set of MDPs, and infinite data. In our experiments, we validate that our insights also hold for finite data and continuous models.} Identifiability strongly depends on the hypothesis set $\mathcal{M}$. However, for learning deep neural network inference models, we do not want to impose any structure on $\mathcal{M}$. Thus, in the following we provide a sufficient identifiability condition that holds for any $\mathcal{M}$.




\begin{proposition}\label{prop:identifiable}
    Consider the setting described in Definition~\ref{def:ambiguity}. For a pair of MDPs $i$ and $j$, we define the identifying state-action pairs as the state-action pairs that satisfy  $\mathcal{R}_i(s, a)\neq \mathcal{R}_j(s, a)$ and/or $\mathcal{P}_i(s'|s, a)\neq\mathcal{P}_j(s'|s, a)$. If for every $i\neq j$ there exists an identifying state-action pair that has positive probability under both $i$ and $j$, i.e., $P_{\mathcal{P}_i, \pi_{\beta}^{i}}(s, a),P_{\mathcal{P}_j, \pi_{\beta}^{j}}(s, a) > 0$, then the data is identifiable.
\end{proposition}
Thus, if the agent has data on identifying state-actions \textit{obtained from different MDPs}, it has the capability to identify which data samples belong to which MDP.
We next categorize several common types of meta-RL problems according to identifiability, as per Proposition~\ref{prop:identifiable}; we will later revisit this dichotomy in our experiments. For our illustration, we assume that in each training MDP, the data collecting policy is approximately optimal (this would be the case when training standard RL agents on each MDP). Let us first consider problems that only differ in the reward. Here, when identifying state-actions (i.e., state-actions with different rewards) in different MDPs do not overlap, we will have an identifiability problem. The sparse reward tasks in Figures \ref{fig:illustration} and \ref{fig:r_ambiguity} are examples of this case -- each agent will visit only its own reward area, resulting in ambiguity. When the rewards are dense, however, it is much more likely that the data is identifiable; common tasks like Half-Cheetah-Vel (cf. Sec.~\ref{sec:experiments}) are examples of this setting. For MDPs that differ in their transitions, a similar argument can be made about whether the identifying state-actions overlap or not. Most studies on online and offline meta-RL to date considered problems with overlapping identifying state-actions, where ambiguity is not an issue. For example, in the Walker environment of \citet{zintgraf2020varibad}, the shape of the agent is varied, which manifests in almost every transition, and a successful agent must walk forward, thus many overlapping state-actions are visited; the Wind domain (cf. Sec.~\ref{sec:experiments}) is another example. Examples of problems with non-overlapping identifying transitions are, for example, peg-in-hole insertion where the hole position varies between tasks, or the Escape-Room domain in Sec.~\ref{sec:experiments}; in such domains we expect ambiguity to be a concern. One can of course imagine combinations and variations of the categories above -- our aim is not to be exhaustive, but to illustrate which OMRL problems are difficult due to ambiguity, and which are not. 

Note that MDP ambiguity is special to offline meta-RL; in online meta-RL, the agent may be driven by the online adapting policy (or guided explicitly) to explore states that reduce its ambiguity. We also emphasize that this problem is not encountered in standard (non-meta) offline RL, as the problem here concerns the \emph{identification of the MDP}, which in standard RL is unique.


How can one collect data to mitigate MDP ambiguity? We present a simple, general modification to the data collection scheme we term \textbf{policy replaying}, which, under mild conditions on the original data collection policies, guarantees that the resulting data will be identifiable. We importantly note that changing the data collection method in-hindsight is not suitable for the offline setting. Therefore, the proposed scheme should be viewed as \textit{a guideline for effective OMRL data collection}. For each MDP, we propose collecting data in the following manner: randomly draw a data collection policy from $\{\pi_{\beta}^{i}\}_{i=1}^{N}$, collect a trajectory following that policy, and repeat.
After this procedure, the new data distributions are all associated with \emph{the same} data collection policy, which we denote $\pi_{r}$.\footnote{Note that even if $\pi_{\beta}^{i}$ are Markov for all $i$, the replaying policy $\pi_{r}$ is not necessarily Markov.} 
\begin{proposition}\label{prop:policy_replaying}
    For every $i\neq j$, denote the set of identifying state-action pairs by $\mathcal{I}_{i,j}$. If for every $i$ and every $j$ exists $(s_{i,j}, a_{i,j})\in\mathcal{I}_{i,j}$ such that $P_{\mathcal{P}_i, \pi_{\beta}^{i}}(s_{i,j}, a_{i,j})>0$, then replacing $\pi_{\beta}^{i}$ with $\pi_{r}$ for all $i$ results in identifiable data.
\end{proposition}

Note that the requirement on identifying states in Proposition \ref{prop:policy_replaying} is minimal -- without it, the original data collecting policies $\pi_{\beta}^{i}$ are useless, as they do not visit any identifying states (e.g., consider the example in Figure \ref{fig:r_ambiguity} with policies that do not visit the reward at all).

When the tasks only differ in their reward function, and the reward functions for the training environments are known, policy replaying can be implemented in hindsight, \textit{without changing the data collection process}. This technique, which we term \textbf{Reward Relabelling (RR)}, is applicable under the offline setting, and described next.
In RR, we replace the rewards in a trajectory from some MDP $i$ in the data with rewards from another randomly chosen MDP $j \neq i$. That is, for each $i\in 1,\dots,N$, we add to the data $K$ trajectories $\hat{\tau}^{i,k},\quad k\in 1,\dots,K$, where $\hat{\tau}^{i,k} = (s_0^{i,k}, a_0^{i,k}, \hat{r}_{1}^{i,k}, s_1^{i,k}\dots, \hat{r}_{H}^{i,k}, s_H^{i,k})$, where the relabelled rewards $\hat{r}$ satisfy $\hat{r}_{t+1}^{i,k} = \mathcal{R}_{j}(s_t^{i,k},a_t^{i,k})$.
Thus, our relabelling effectively runs $\pi_{\beta}^{i}$ on MDP $j$, which is equivalent to performing policy replaying (in hindsight). We remark that the assumption on known reward (during training) is mild, as the reward is the practitioner's method of specifying the task goal, which is typically known~\citep{gu2017deep, schwab2019simultaneously, schoettler2020meta}; this assumption is also satisfied in all meta-RL studies to date.

\textbf{\acronym}: we refer to the \acronym\ algorithm as the combination of the policy replaying/RR techniques and off-policy RL applied to state-relabelled trajectories. We provide pseudo-code in the supplementary material (see Appendix~\ref{appendix:pseudo_code}).
\section{Related Work}
We focus on meta-RL -- quickly learning to solve RL problems.
Gradient based approaches to meta-RL seek policy parameters that can be updated to the current task with a few gradient steps~\citep{finn2017model,grant2018recasting,rothfuss2018promp,clavera2018model}. These are essentially online methods, and several studies investigated learning of structured exploration strategies in this setting~\citep{gupta2018meta,rothfuss2018promp,stadie2018some}.
Memory-based meta-RL, on the other hand, map the observed history in a task $h_{:t}$ to an action~\citep{duan2016rl,wang2016learning}. These methods effectively treat the problem as a POMDP, and learn a memory based controller for it. 

The connection between meta-learning and Bayesian methods, and between meta-RL and Bayesian RL in particular, has been investigated in a series of recent papers~\citep{lee2018bayesian,humplik2019meta,ortega2019meta,zintgraf2020varibad}, and our work closely follows these ideas. In particular, these works elucidate the difference between Thompson-sampling based strategies, such as PEARL~\citep{rakelly2019efficient}, and Bayes-optimal policies, such as VariBAD, and suggest to estimate the BAMDP belief using the latent state of deep generative models. \emph{Our contribution is an extension of these ideas to the offline RL setting}, which to the best of our knowledge is novel. Technically, the VariBAD algorithm in~\cite{zintgraf2020varibad} is limited to on-policy RL, and the off-policy method in~\cite{humplik2019meta} requires specific task descriptors during learning, while VariBAD, which our work is based on, does not. One can also learn neural belief models using contrastive learning~\cite{guo2018neural}; our methods and identifiability discussion apply to this case as well.

\textbf{Concurrently and independently with our work}, \citet{li2020multitask} proposed an offline meta-RL algorithm that combines BCQ~\citep{fujimoto2018off} with a task inference module. 
Interestingly, \citet{li2020multitask} also describe a problem similar to MDP ambiguity, and resolve it using a technique similar to reward relabelling. However, their approach does not take into account the task uncertainty, and cannot plan actions that actively explore to reduce this uncertainty -- this is a form of Thompson sampling, where a task-conditional policy reacts to the task inference (see Figure 1 in \citealt{li2020multitask}). \textbf{Our work is the first to tackle offline meta-learning of Bayes-optimal exploration}. In addition, we demonstrate the first offline results on sparse reward tasks, which, compared to the dense reward tasks in \citet{li2020multitask}, require a significantly more complicated solution than Thompson sampling (see experiments section). We achieve this by building on BRL theory, which both optimizes for Bayes-optimality and results in a much simpler algorithm.
Recent work on meta Q-learning~\citep{fakoor2019meta} also does not incorporate task uncertainty, and thus 
cannot be Bayes-optimal. The very recent work of \citet{mitchell2020offline} considers a different offline meta-RL setting, where an offline dataset from the test environment is available. 

Classical works on BRL are comprehensively surveyed by~\citet{ghavamzadeh2016bayesian}. Our work, in comparison, allows training scalable deep BRL policies. Finally, there is growing interest in offline deep RL~\citep{sarafian2018safe,levine2020offline}. Most recent work focus on how to avoid actions that were not sampled enough in the data. In our experiments, a state-of-the-art method of this flavor led to minor improvements, though future offline RL developments may possibly benefit OMRL too.
\section{Experiments}\label{sec:experiments}
In our experiments, we aim to demonstrate:
(1) Learning approximately Bayes-optimal policies in the offline setting; and (2) The severity of MDP ambiguity, and the effectiveness of our proposed resolutions. In the supplementary material, we also report that our off-policy method improves meta-RL performance in the online setting. 
\begin{figure*}[t]
    \centering
    \begin{subfigure}{0.2\textwidth}
        \centering
        \includegraphics[width=\linewidth]{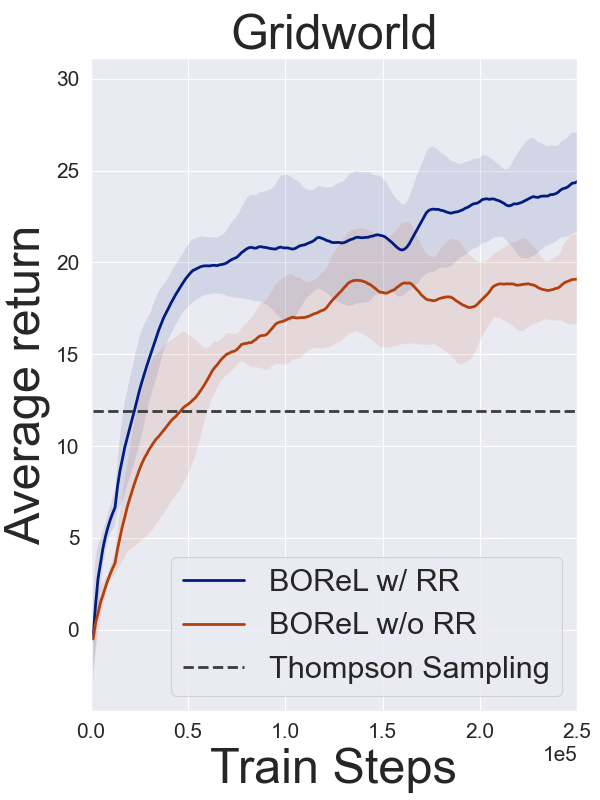} 
        \label{fig:offline_res_gridworld}
    \end{subfigure}\hfill
    \begin{subfigure}{0.2\textwidth}
        \centering
        \includegraphics[width=\linewidth]{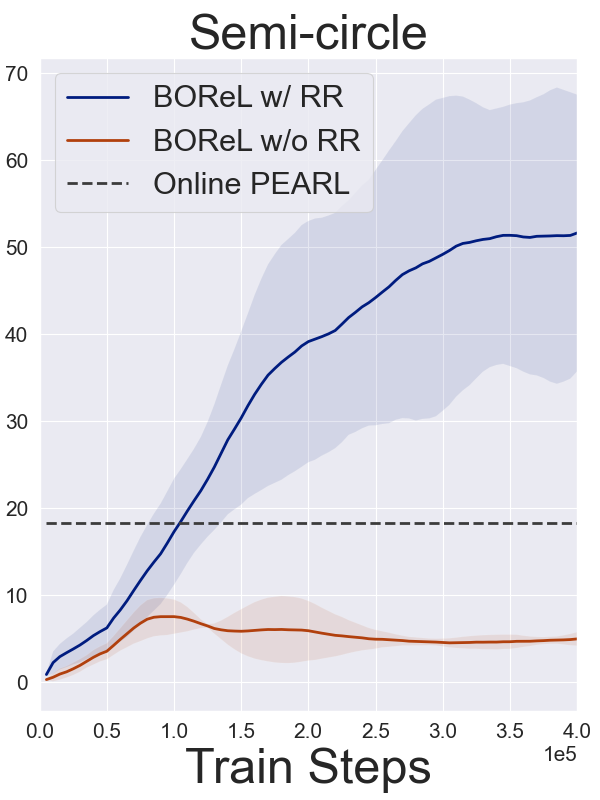} 
        \label{fig:offline_res_halfcircle}
    \end{subfigure}\hfill
    \begin{subfigure}{0.2\textwidth}
        \centering
        \includegraphics[width=\linewidth]{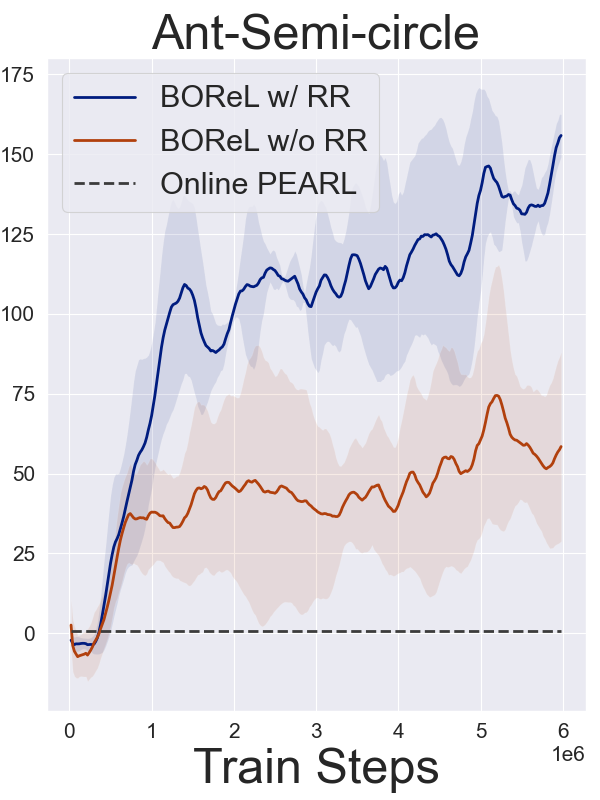} 
        \label{fig:offline_res_ant_semicircle}
    \end{subfigure}\hfill
    \begin{subfigure}{0.2\textwidth}
        \centering
        \includegraphics[width=\linewidth]{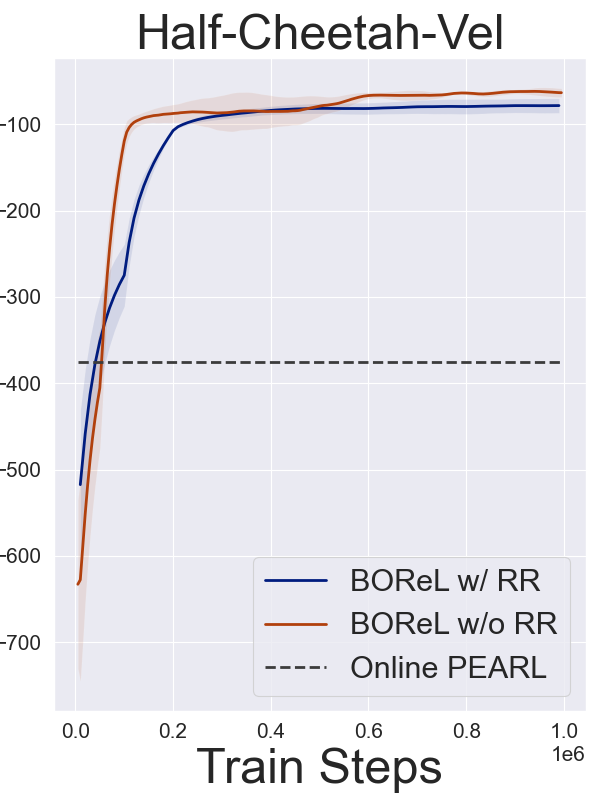} 
        \label{fig:offline_res_cheetah_vel}
    \end{subfigure}\hfill
    \begin{subfigure}{0.2\textwidth}
        \centering
        \includegraphics[width=\linewidth]{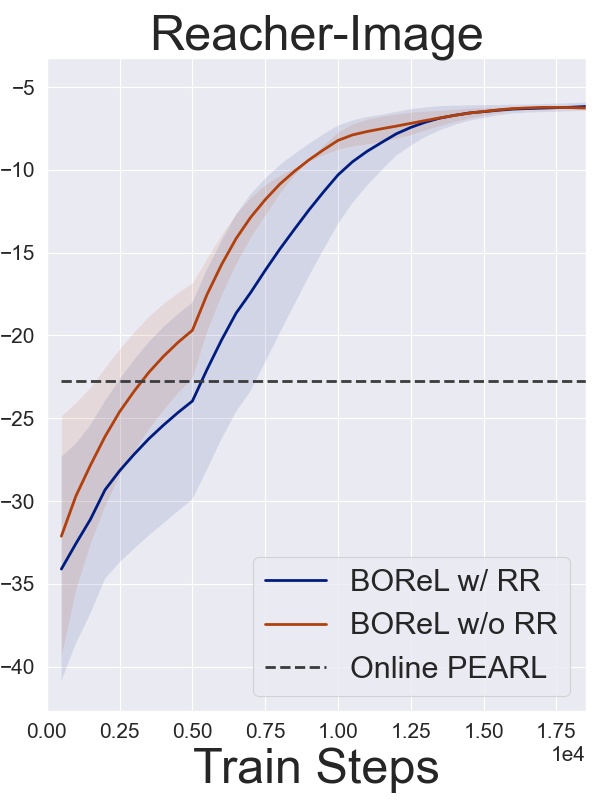} 
        \label{fig:offline_res_reacher}
    \end{subfigure}
    \vspace{-1em}
    \caption{Offline performance on domains with varying rewards. 
    We compare \acronym{} with and without reward relabeling (blue and red, respectively) with Thompson sampling baselines -- calculated exactly in Gridworld, and using online PEARL for the other domains. Full training curves for PEARL are reported in the supplementary; here we plot only the best performance.}
    \label{fig:offline_res}
    \vspace{-1em}
\end{figure*}

Answering (1) is difficult because the Bayes-optimal policy is generally intractable, and because our results crucially depend on the available data. However, in deterministic domains with a single sparse reward, the optimal solution amounts to `search all possible goal locations as efficiently as possible, and stay at goal once found; in subsequent episodes, move directly to goal'. We therefore chose domains where this behavior can be identified qualitatively. Quantitatively, we compare our offline results with online methods based on Thompson sampling, which are not Bayes-optimal, and aim to show that the performance improvement due to being approximately Bayes-optimal gives an advantage, \textit{even under the offline data} restriction.
Our code is available online at \url{https://github.com/Rondorf/BOReL}.
\paragraph{Domains and evaluation metric:} we evaluate learning to explore efficiently in a diverse set of domains: (1) A discrete $5\times 5$ Gridworld  \citep{zintgraf2020varibad}; (2) Semi-circle -- a continuous point robot where a sparse reward is located somewhere on a semi-circle (see Figure~\ref{fig:illustration}); (3) Ant-Semi-circle -- a challenging modification of the popular Ant-Goal task \citep{fakoor2019meta} to a sparse reward setting similar to the semi-circle task above (see Figure \ref{fig:ant_trajs}); 
(4) Half-Cheetah-Vel \citep{finn2017model}, a popular high-dimensional control domain with dense rewards; (5) Reacher-Image -- 2-link robot reaching an unseen target located somewhere on a quarter circle, with image observations and dense rewards (see Appendix~\ref{appendix:env_description}); (6) Wind -- a point robot navigating to a fixed goal in the presence of varying wind; and (7) Escape-Room -- a point robot that needs to escape a circular room where the only opening is somewhere on the semi-circle (full details in Appendix~\ref{appendix:env_description}). These domains portray both discrete (1) and continuous (2-7) dynamics, and environments that differ either in the rewards (1-5) or transitions (6-7). Domains (3), (4) and (5) are high-dimensional, and the navigation problems (1-3, 7) require non-trivial exploration behavior to quickly identify the task. Importantly, relating to the MDP ambiguity discussion in Sec.~\ref{sec:indentifiability}, domains (1-3, 7) have non-overlapping identifying states; here we expect MDP ambiguity to be a problem. On the other hand, in domains (4-6) the identifying states are expected to overlap, as the rewards/transition differences are dense. 
To evaluate performance, we measure average reward in the first $2$ episodes on unseen tasks -- this is where efficient exploration makes a critical difference.\footnote{For Gridworld, we measure average reward in the first $4$ episodes, and for Wind, only the first episode reward is measured.} In the supplementary, we report results for more evaluation episodes. 


\begin{figure}
    \centering
    \begin{subfigure}{0.23\textwidth}
        \centering
        \includegraphics[width=\linewidth]{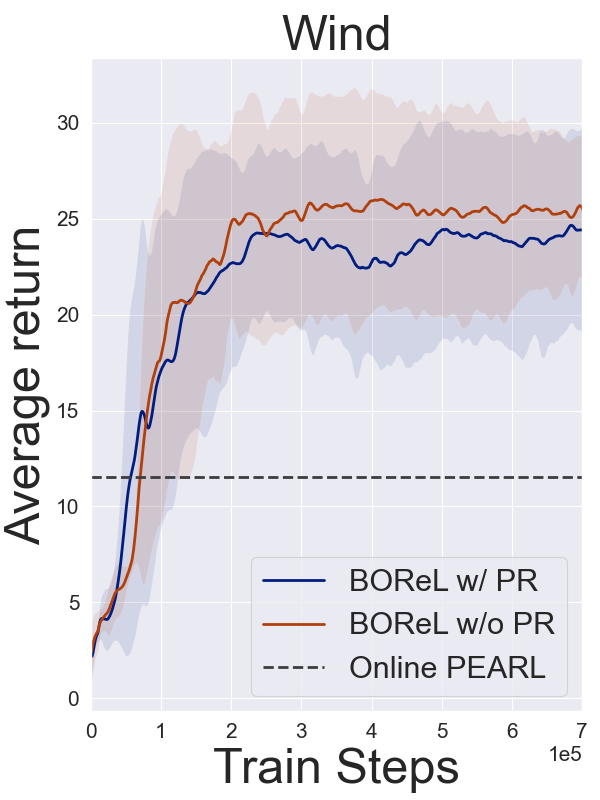}
        \label{fig:offline_res_wind}
    \end{subfigure}\hfill
    \begin{subfigure}{0.23\textwidth}
        \centering
        \includegraphics[width=\linewidth]{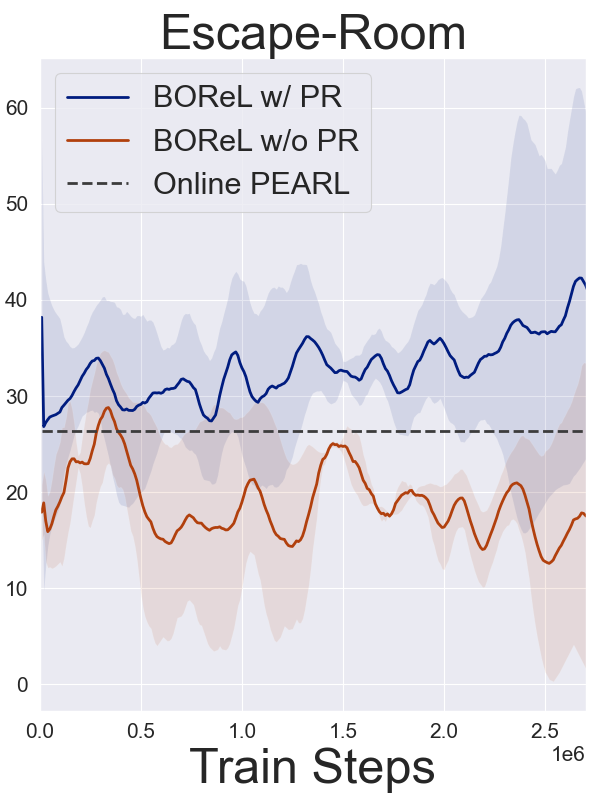} 
        \label{fig:offline_res_escape_room}
    \end{subfigure}
    \vspace{-1em}
    \caption{Offline performance on domains with varying transitions. 
    We compare \acronym{} instantiated with and without policy replaying (blue and red, respectively) with online PEARL.}
    \label{fig:offline_res_trnasitions}
    \vspace{-1.5em}
\end{figure}

\vspace{-1em}
\paragraph{Data collection and organization:} For data collection, we used off-the-shelf DQN (Gridworld) and SAC (continuous domains) implementations.\footnote{Note that collecting offline data for each domain requires to first successfully train a large number of `standard' RL agents, which can be demanding; we will make our data available publicly.}
To study the effect of data diversity, we diversified the offline dataset by modifying the initial state distribution $P_{init}$ to either (1) uniform over a large region, (2) uniform over a restricted region, or (3) fixed to a single position. At meta-test time, only the single fixed position is used.
The tasks are episodic, but we want the agent to maintain its belief between episodes, so that it can continually improve performance (see Figure~\ref{fig:illustration}). We follow \citet{zintgraf2020varibad}, and aggregate $k$ consecutive episodes of length $H$ to a long trajectory of length $k\times H$, and we do not reset the hidden state in the VAE recurrent neural network after episode termination. For reward relabelling, we replace either the first or last $k/2$ trajectories with trajectories from a randomly chosen MDP, and relabel their rewards. For policy replay we replace trajectories by sampling a new trajectory using the trained RL policy of another MDP.
Technically, network architectures and hyperparameters were chosen similarly to \citet{zintgraf2020varibad}, as detailed in the supplementary.
\vspace{-1em}
\paragraph{Main Results:}
In Figure \ref{fig:offline_res} we compare our offline algorithm with Thompson sampling based methods, and also with an ablation of the reward relabelling method. For Gridworld, the Thompson sampling method is computed exactly, while for the continuous environments, we use online PEARL~\citep{rakelly2019efficient} -- a strong baseline that is \textit{not affected by our offline data  limitation}. In particular, this baseline is stronger than the offline meta-RL algorithm of \citet{li2020multitask}. For these results the uniform initial state distribution was used to collect data. Note that \textbf{we significantly outperform Thompson sampling based methods, demonstrating our claim of learning non-trivial exploration from offline data}. These results are further explained qualitatively by observing the exploration behavior of our learned agents. In Figure \ref{fig:illustration} and in Figure~\ref{fig:ant_trajs}, we visualize the trajectories of the trained agents in the Semi-circle and Ant-Semi-circle domains, respectively.\footnote{Video is provided: \url{https://youtu.be/6Swg55ZYOU4}} 
An approximately Bayes-optimal behavior is evident: in the first episode, the agents search for the goal along the semi-circle, and in the second episode, the agents maximize reward by moving directly towards the already found goal. Similar behaviors for Gridworld and Escape-Room are reported in Appendix~\ref{appendix:learned_belief}. In contrast, a Thompson sampling based agent will never display such search behavior, as \textit{it does not plan} to proactively reduce uncertainty. Instead, such an agent will randomly choose an un-visited possible goal at each episode and directly navigate towards it (cf. Figure 1 in \citealt{li2020multitask}). We further emphasize that the approximately Bayes-optimal search behavior is very different from the training data, in which the agents learned to reach specific goals. 

\begin{figure}
    \centering
    \includegraphics[width=0.95\linewidth]{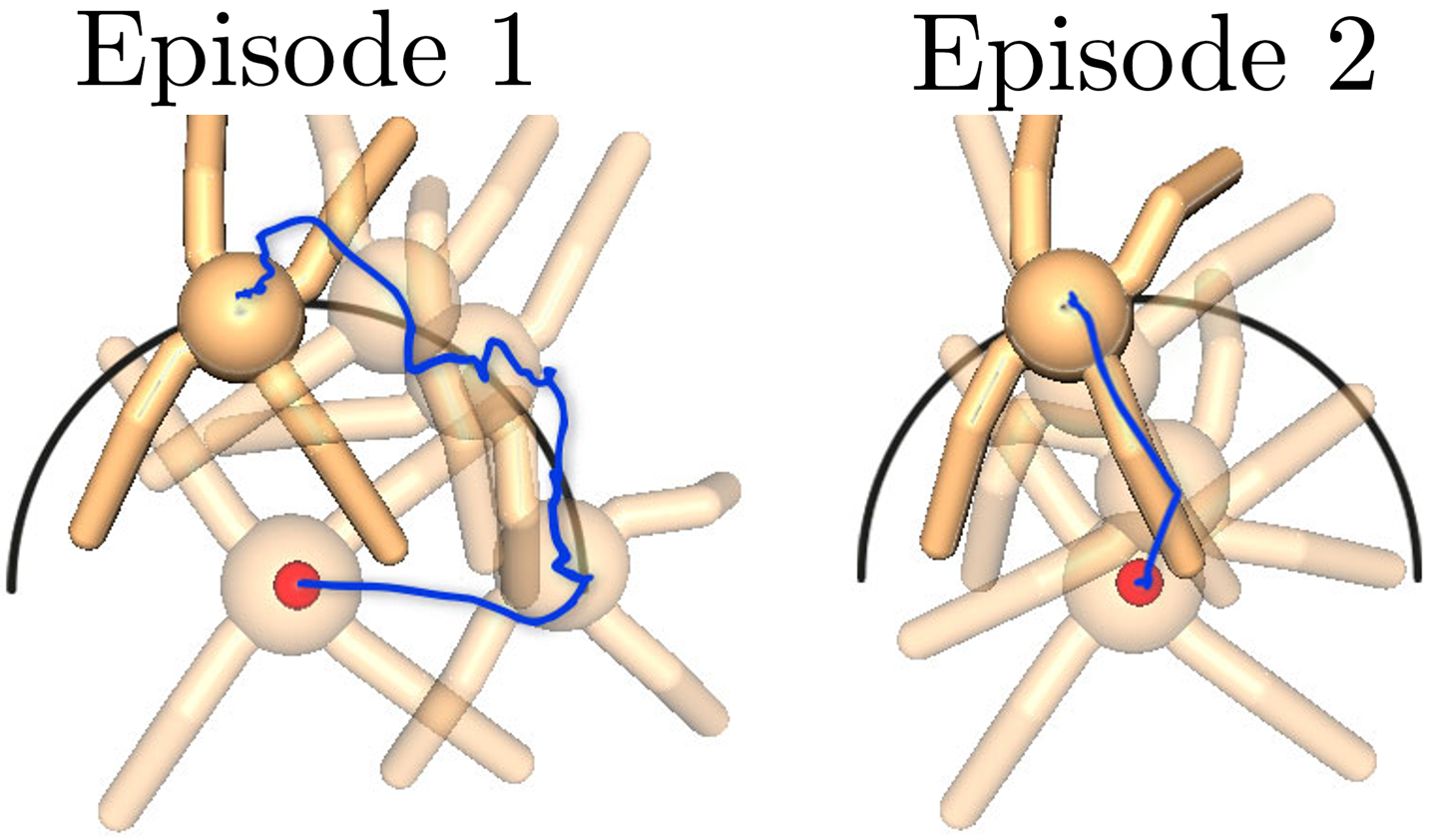}
    \caption{Ant-Semi-circle: trajectories from trained policy on a new goal. 
    Note that in the first episode, the ant searches for the goal, and in the second one it directly moves toward the goal it has previously found. This search behavior is different from the goal-reaching behaviors that dominate the training data.}
    \label{fig:ant_trajs}
    \vspace{-1.5em}
\end{figure}

Our results also signify the severity of MDP ambiguity, and align with the theory in Sec.~\ref{sec:indentifiability}. In domains with non-overlapping identifying states (1-3, 7), as expected, performance without policy replaying (RR) is poor, while in domains with overlapping identifying states policy replaying has little effect. In Figure~\ref{fig:belief_halfcircle} in the supplementary, we provide further insight into these results, by plotting the belief update during the episode rollout for Semi-circle: the belief starts as uniform on the semi-circle, and narrows in on the target as the agent explores the semi-circle. With reward relabelling ablated, however, we show that the belief does not update correctly, and the agent believes the reward is at the point it first visited on the semi-circle.



\paragraph{Data Quality Ablative Study:} To evaluate the dependency of our method on the offline data quality, we report results for the $3$ different data collection strategies described above (see supplementary for more details), 
summarized in Table~\ref{wrap-tab:1}. As expected, data diversity is instrumental to offline training. However, as we qualitatively show in Figure~\ref{fig:visualize_trajs_initial_dist} in the supplementary, even on the low-diversity datasets, our agents learned non-trivial exploration strategies that search for the goal. This is especially remarkable for the fixed-distribution dataset, where it is unlikely that any training trajectory traveled along the semi-circle.

One may ask whether OMRL presents the same challenge as standard offline RL, and whether recent offline RL advances can mitigate the dependency on data diversity. To investigate this, we also compare our method with a variant that uses CQL \citep{kumar2020conservative} -- a state-of-the-art offline RL method -- to train the critic network of the meta-RL agent. 
Interestingly, while CQL improved results (Table \ref{wrap-tab:1}), the data diversity is much more significant. Together with our results on MDP ambiguity, our investigation highlights the particular challenges of the OMRL problem.
\begin{table}
\caption{Average return in Ant-Semi-circle for different initial state distributions during offline data collection: \textbf{Uniform} distribution, uniform distribution excluding states on the semi-circle (\textbf{Excluding s.c.}), and fixed initial position (\textbf{Fixed}).} 
\centering
\begin{tabular}{ccc}\\\toprule  
& Ours & w/ CQL \\  \midrule
Uniform &171.8 $\pm$ 7.0 & 176.0 $\pm$ 10.2 \\  \midrule
Excluding s.c. & 102.8 $\pm$ 32.7& 116.6 $\pm$ 19.9 \\  \midrule
Fixed &99.2 $\pm$ 27.4 & 112.4 $\pm$ 31.3 \\  \bottomrule
\end{tabular}
\label{wrap-tab:1}
\vspace{-1em}
\end{table}



\section{Conclusion and Future Work}
We presented the first offline meta-RL algorithm that is approximately Bayes-optimal, allowing to solve problems where efficient exploration is crucial. The connection between Bayesian RL and meta learning allows to reduce the problem to offline RL on belief-augmented states. However, learning a neural belief update from offline data is prone to the MDP ambiguity problem. We formalized the problem, and proposed a simple data collection protocol that guarantees identifiability. In the particular case of tasks that differ in their rewards, our protocol can be implemented in hindsight, for arbitrarily offline data. Our results show that this solution is effective on several challenging domains.

It is intriguing whether other techniques can mitigate MDP ambiguity. For example, designing data collection policies that diversify the data or injecting prior knowledge by controlling the hypothesis set of the neural belief update. Extending our investigation to general POMDPs is another interesting direction.


\bibliography{references.bib}
\bibliographystyle{icml2021}


\appendix
\onecolumn

\section{Propositions Proofs}
For ease of reading, we copy here the propositions from the main text. 

\textbf{Proposition 1.}\textit{
Let $\tau=s_0,a_0,r_1,s_1\dots, r_H, s_H$ denote a random trajectory from a fixed history dependent policy $\pi$, generated according to the following process. First, MDP parameters $\mathcal{R},\mathcal{P}$ are drawn from the prior $p(\mathcal{R},\mathcal{P})$. Then, the state trajectory is generated according to  $s_0\sim P_{init}$, $a_t\sim \pi(\cdot | s_0,a_0,r_1,\dots,s_t)$, $s_{t+1}\sim \mathcal{P}(\cdot|s_t,a_t)$ and $r_{t+1}\sim \mathcal{R}(s_t,a_t)$. Let $b_t$ denote the posterior belief at time t, $b_t = P(\mathcal{R},\mathcal{P} | s_0,a_0,r_1,\dots,s_t)$. Then $$P(s_{t+1}|s_0,a_0,r_1,\dots,r_t, s_t, a_t)=\mathbb{E}_{\mathcal{R},\mathcal{P}\sim b_t} \mathcal{P}(s_{t+1}|s_t, a_t), \text{ and,}$$
$$P(r_{t+1}|s_0,a_0,r_1,\dots,s_t, a_t)=\mathbb{E}_{\mathcal{R},\mathcal{P}\sim b_t} \mathcal{R}(r_{t+1}|s_t, a_t).$$}

\begin{proof}
For the transitions, we have that,
\begin{equation*}
\begin{split}
    P(s_{t+1}|s_0,a_0,r_0,\!\dots\!,r_t,s_t,a_t)\! &=\int{P(s_{t+1}, \mathcal{R}, \mathcal{P}|s_0,a_0,r_0,\!\dots\!,r_t,s_t,a_t)d\mathcal{R}d\mathcal{P}} \\ &=\int{P(s_{t+1}| \mathcal{R}, \mathcal{P}, s_0,a_0,r_0,\!\dots\!,r_t,s_t,a_t) P(\mathcal{R}, \mathcal{P}|s_0,a_0,r_0,\!\dots\!,r_t,s_t,a_t)d\mathcal{R}d\mathcal{P}} \\ &= \!\mathbb{E}_{\mathcal{R},\mathcal{P}}\!\left[\left. P(s_{t+1}|\mathcal{R},\mathcal{P},s_0,a_0,r_0,\dots,r_t,s_t,a_t)\right| s_0,a_0,r_0,\dots,r_t,s_t,a_t\right] \\
    &= \mathbb{E}_{\mathcal{R},\mathcal{P}}\left[\left. \mathcal{P}(s_{t+1}|s_t,a_t)\right| s_0,a_0,r_0,\dots,r_t,s_t,a_t\right] \\
    &= \mathbb{E}_{\mathcal{R},\mathcal{P}\sim b_t} \mathcal{P}(s_{t+1}|s_t, a_t).
\end{split}
\end{equation*}
The proof for the rewards proceeds similarly.
\end{proof}

\subsection*{Extended Definitions and Proofs for Section 4}
For the proofs of identifiability, we start by elaborating the formal definition of our setting.
For simplicity, we assume that the MDPs $\{\mathcal{R}_i,\mathcal{P}_i\}_{i=1}^{N}$ are defined over finite state-action spaces ($\lvert\mathcal{S}\rvert, \lvert\mathcal{A}\rvert<\infty$).
For every $i=1,\ldots,N$, let $\pi_{\beta}^{i}$ be a general stationary, stochastic, history-dependent policy\footnote{We consider stationary policies for notation simplicity, although similar analysis can be made for non-stationary policies.}. The initial state distribution $P_{init}$ is the same across all MDPs.\footnote{The idea of policy replaying can be extended to MDP with different initial state distributions by randomly selecting the state distribution along with the policy. For simplicity, we do not consider this case, although a similar analysis holds for it.}

We assume that data is collected from trajectories of length at most $T_{\max}$. This is a convenient assumption that holds in every practical scenario, and allows us to side step issues of defining visitation frequencies when $t\to\infty$. 

For some $0\leq t \leq T_{\max}$, denote by $P_{\mathcal{P}_i, \pi_\beta^{i}, t}(s, a) = P_{\mathcal{P}_i, \pi_\beta^{i}}(s_t=s, a_t=a)$ the probability of visiting the state-action pair $(s,a)$ at time $t$ by running the policy $\pi_{\beta}^{i}$ on MDP with transition function $\mathcal{P}_i$ and initial state distribution $P_{init}$. Now, we define: 
    \[
        P_{\mathcal{P}_i, \pi_\beta}(s,a) = P_{\mathcal{P}_i, \pi_\beta^{i}}\Big(\bigcup_{t\in\{0,\ldots,T_{\max}\}}\{s_t=s, a_t=a\}\Big),
    \]
that is, $P_{\mathcal{P}_i, \pi_\beta}(s,a)$ is the probability of observing state-action $(s,a)$ in the data from MDP $i$. Similarly, we define
\begin{equation*}
    P_{\mathcal{P}_i, \mathcal{R}_i,\pi_\beta}(s,a,r,s')=P_{\mathcal{P}_i, \pi_\beta}(s,a)\mathcal{P}_{i}(s'|s,a)P_{\mathcal{R}_i}(r|s,a),
\end{equation*}
the probability of observing the tuple $(s,a,r,s')$ in the data from MDP $i$.

A trajectory from the replay policy $\pi_r$ in MDP $i$ is generated as follows. Let $x$ be a discrete random variable defined on $1,\dots,N$ with probability $P_x(\cdot)$ that satisfies $P_{x}(x=k)>0$ for every $k=1,\ldots,N$. First, we draw $x$.
Then, we sample a trajectory from MDP $i$ using policy $\pi_{\beta}^{x}$.


\textbf{Proposition 2.}\textit{
    Consider the setting described in Definition~\ref{def:ambiguity}. For a pair of MDPs $i$ and $j$, we define the identifying state-action pairs as the state-action pairs that satisfy $\mathcal{R}_i(s, a)\neq \mathcal{R}_j(s, a)$ and/or $\mathcal{P}_i(s'|s, a)\neq\mathcal{P}_j(s'|s, a)$. If for every $i\neq j$ there exists an identifying state-action pair that has positive probability under both $i$ and $j$, i.e., $P_{\mathcal{P}_i, \pi_{\beta}^{i}}(s, a),P_{\mathcal{P}_j, \pi_{\beta}^{j}}(s, a) > 0$, then the data is identifiable.
}

Before we prove Proposition 2, we present the following lemma, which will be used later in the proof.
    \begin{Lem}\label{lem:lem1}
    Consider a pair of MDPs $(\mathcal{R}, \mathcal{P})$ and $(\mathcal{R}', \mathcal{P}')$, and two policies $\pi$ and $\pi'$. If there exists an identifying state-action pair of the MDPs $(\bar{s},\bar{a})$ that has positive probability under both $(\mathcal{P}, \pi)$ and $(\mathcal{P}', \pi')$, i.e., $P_{\mathcal{P}, \pi}(\bar{s}, \bar{a}),P_{\mathcal{P}', \pi'}(\bar{s}, \bar{a}) > 0$, then $P_{\mathcal{R},\mathcal{P},\pi}(s,a,r,s')\neq P_{\mathcal{R}',\mathcal{P}',\pi'}(s,a,r,s')$.
    \end{Lem}
    \begin{proof}
        Assume to the contrary that $P_{\mathcal{R},\mathcal{P},\pi}(s,a,r,s')=P_{\mathcal{R}',\mathcal{P}',\pi'}(s,a,r,s')$. Marginalizing over $r$ and $s'$, we obtain:
        \begin{align*}
            \sum_{r,s'}{P_{\mathcal{R},\mathcal{P},\pi}(s,a,r,s')} &= \sum_{r,s'}{P_{\mathcal{R},\mathcal{P}',\pi'}(s,a,r,s')} \\ P_{\mathcal{P},\pi}(s, a) &= P_{\mathcal{P}',\pi'}(s, a), \quad\forall (s,a).
        \end{align*}
        Specifically, we have $P_{\mathcal{P},\pi}(\bar{s}, \bar{a}) = P_{\mathcal{P}',\pi'}(\bar{s}, \bar{a})$. Since $P_{\mathcal{R},\mathcal{P},\pi}(\bar{s},\bar{a},r,s') = P_{\mathcal{R},\mathcal{P}}(r,s'|\bar{s},\bar{a})P_{\mathcal{P}, \pi}(\bar{s},\bar{a})$ for every $r$ and $s'$, and $P_{\mathcal{P},\pi}(\bar{s}, \bar{a}) = P_{\mathcal{P}',\pi'}(\bar{s}, \bar{a})>0$, it holds that $P_{\mathcal{R},\mathcal{P}}(r,s'|\bar{s},\bar{a})=P_{\mathcal{R}',\mathcal{P}'}(r,s'|\bar{s},\bar{a})$ for every $r$ and $s'$. By marginalizing over $s'$ we get that
        \begin{align*}
            \sum_{s'}{P_{\mathcal{R},\mathcal{P}}(r,s'|\bar{s},\bar{a})} &= \sum_{s'}{P_{\mathcal{R}',\mathcal{P}'}(r,s'|\bar{s},\bar{a})} \\ P_{\mathcal{R}}(r|\bar{s},\bar{a}) &= P_{\mathcal{R}'}(r|\bar{s},\bar{a}).
        \end{align*}
        Similarly, by marginalizing over $r$, we get $\mathcal{P}_{i}(s'|\bar{s},\bar{a})=\mathcal{P}_{j}(s'|\bar{s},\bar{a})$. Overall, both reward and transition function do not differ in $(\bar{s}, \bar{a})$, which contradicts the fact that $(\bar{s}, \bar{a})$ is an identifying state-action pair.
    \end{proof}

We now prove Proposition 2.

\begin{proof}
    Consider some $i\neq j$. Let $(s_{i,j}, a_{i,j})$ be an identifying state-action pair that has positive probability under both $i$ and $j$. Assume to the contrary that there exists an MDP $\{\mathcal{R}, \mathcal{P}\}\in\mathcal{M}$ and two policies $\pi$ and $\pi'$ such that $P_{\mathcal{R}_i, \mathcal{P}_i, \pi_{\beta}^{i}}(s, a, r, s') = P_{\mathcal{R}, \mathcal{P}, \pi}(s, a, r, s')$ and $P_{\mathcal{R}_j, \mathcal{P}_j, \pi_{\beta}^{j}}(s, a, r, s') = P_{\mathcal{R}, \mathcal{P}, \pi'}(s, a, r, s')$. 
    
    Since $(s_{i,j}, a_{i,j})$ has positive probability under $(\mathcal{P}_i, \pi_{\beta}^{i})$ and $P_{\mathcal{R}_i, \mathcal{P}_i, \pi_{\beta}^{i}}(s, a, r, s') = P_{\mathcal{R}, \mathcal{P}, \pi}(s, a, r, s')$, then $(s_{i,j}, a_{i,j})$ must also have positive probability under $(\mathcal{P}, \pi)$ (otherwise, there are $r$ and $s'$ for which $P_{\mathcal{R}_i, \mathcal{P}_i, \pi_{\beta}^{i}}(s_{i,j}, a_{i,j}, r, s')> 0$, while $P_{\mathcal{R}, \mathcal{P}, \pi}(s_{i, j}, a_{i,j}, r, s')=0$). Now, since $(s_{i,j}, a_{i,j})$ has positive probability under both $(\mathcal{P}_i, \pi_{\beta}^{i})$ and $(\mathcal{P}, \pi)$, and $P_{\mathcal{R}_i, \mathcal{P}_i, \pi_{\beta}^{i}}(s, a, r, s') = P_{\mathcal{R}, \mathcal{P}, \pi}(s, a, r, s')$, according to Lemma~\ref{lem:lem1}, it cannot be an identifying state-action pair of $(\mathcal{R}_i, \mathcal{P}_i)$ and $(\mathcal{R}, \mathcal{P})$. Therefore, the MDP $\{\mathcal{R}, \mathcal{P}\}$ must satisfy $\mathcal{P}(\cdot|s_{i,j}, a_{i,j})=\mathcal{P}_i(\cdot|s_{i,j}, a_{i,j})$ and $\mathcal{R}(s_{i,j}, a_{i,j})=\mathcal{R}_i(s_{i,j}, a_{i,j})$.
    
    The same argument can be made for $(\mathcal{R}_j, \mathcal{P}_j, \pi_{\beta}^{j})$ and $(\mathcal{R}, \mathcal{P}, \pi')$, resulting in $\mathcal{P}(\cdot|s_{i,j}, a_{i,j})=\mathcal{P}_j(\cdot|s_{i,j}, a_{i,j})$ and $\mathcal{R}(s_{i,j}, a_{i,j})=\mathcal{R}_j(s_{i,j}, a_{i,j})$. Overall, we get $\mathcal{P}_i(\cdot|s_{i,j}, a_{i,j})=\mathcal{P}(\cdot|s_{i,j}, a_{i,j})=\mathcal{P}_j(\cdot|s_{i,j}, a_{i,j})$ and $\mathcal{R}_i(s_{i,j}, a_{i,j})=\mathcal{R}(s_{i,j}, a_{i,j})=\mathcal{R}_j(s_{i,j}, a_{i,j})$, which is a contradiction, as $(s_{i,j}, a_{i,j})$ is an identifying state-action pair of MDPs $i$ and $j$.
\end{proof}

\textbf{Proposition 3.}\textit{
    For every $i\neq j$, denote the set of identifying state-action pairs by $\mathcal{I}_{i,j}$. If for every $i$ and every $j$ exists $(s_{i,j}, a_{i,j})\in\mathcal{I}_{i,j}$ such that $P_{\mathcal{P}_i, \pi_{\beta}^{i}}(s_{i,j}, a_{i,j})>0$, then replacing $\pi_{\beta}^{i}$ with $\pi_{r}$ for all $i$ results in identifiable data.
}
    
\begin{proof}
    
    Consider some $i\neq j$. We observe that by the construction of $\pi_r$, for every $(s,a)$ pair that satisfies $P_{\mathcal{P}_i, \pi_{\beta}^i}(s,a)>0$, we also have $P_{\mathcal{P}_i, \pi_{r}}(s,a)>0$. In particular, we have $P_{\mathcal{P}_i, \pi_{r}}(s_{i,j},a_{i,j})>0$. 
    
    We will show that either $(s_{i,j}, a_{i,j})$ also has positive probability under $(\mathcal{P}_j, \pi_r)$ or there must exist some other state-action pair that has positive probability under both $(\mathcal{P}_i, \pi_r)$ and $(\mathcal{P}_j,\pi_r)$. This, according to Proposition~\ref{prop:identifiable}, will result in identifiability of the data.
    
    We define the following sets of state-action pairs:
    \begin{align*}
        &\Sigma_t^{i} = \left\{(s,a): P_{\mathcal{P}_i, \pi_r,t}(s,a)>0\right\}, \quad t=0,1,\ldots,T_{\max}, \\ &\Sigma_t^{i,j} = \left\{(s,a): P_{\mathcal{P}_i, \pi_r,t}(s,a) = P_{\mathcal{P}_j, \pi_r,t}(s,a)>0\right\}, \quad t=0,1,\ldots,T_{\max}.
    \end{align*}
    Note that $\Sigma_{0}^{i}=\Sigma_{0}^{i,j}$, as the initial state distribution $P_{init}$ and $\pi_r$ are fixed across all MDPs. 
    
    First, consider the case where for every $t=0,1,\ldots,T_{\max}$ we have $\Sigma_t^i=\Sigma_t^{i,j}$.
    Given that $(s_{i,j}, a_{i,j})$ has positive probability under $(\mathcal{P}_i,\pi_r)$, there exists some $t$ for which $(s_{i,j}, a_{i,j})\in\mathcal{I}_{i,j}\cap\Sigma_{t}^{i}$. Since $\Sigma_{t}^{i}=\Sigma_{t}^{i,j}$, we have $(s_{i,j}, a_{i,j})\in\mathcal{I}_{i,j}\cap\Sigma_{t}^{i,j}$, which means $(s_{i,j}, a_{i,j})$ also has positive probability under $(\mathcal{P}_j, \pi_r)$.
    
    Next, consider the case where there exists some $t\in\{1,\ldots,T_{\max}\}$ for which $\Sigma_{t}^{i}\neq\Sigma_{t}^{i,j}$ and let $\hat{t}=\min\{t: \Sigma_{t}^{i}\neq\Sigma_t^{i,j}\}$.
    Note that $\hat{t}>0$, since we have already shown that $\Sigma_{0}^{i}=\Sigma_{0}^{i,j}$. Thus, for every $t<\hat{t}$ we have $\Sigma_{t}^{i}=\Sigma_t^{i,j}=\Sigma_t^{j}$, and for $\hat{t}$ it holds that $P_{\mathcal{P}_i, \pi_r, \hat{t}}(s, a) \neq P_{\mathcal{P}_j, \pi_r, \hat{t}}(s, a)$. If there exits a $t'<\hat{t}-1$ and $(s,a)\in\Sigma_{t'}^{i}$ such that $\mathcal{P}_i(\cdot|s,a)\neq\mathcal{P}_j(\cdot|s,a)$, then we are done as $\Sigma_{t'}^{i}=\Sigma_{t'}^{i,j}$, which means that $(s,a)$ is an identifying state-action pair that has positive probability under both $(\mathcal{P}_i,\pi_r)$ and $(\mathcal{P}_j, \pi_r)$. Therefore, consider the case where for every $t<\hat{t}-1$ and every $(s,a)\in\Sigma_{t}^{i}$ we have $\mathcal{P}_i(\cdot|s,a)=\mathcal{P}_j(\cdot|s,a)$. We will show that there exists $(s,a)\in \Sigma_{\hat{t}-1}^{i}$ such that $\mathcal{P}_i(\cdot|s,a)\neq \mathcal{P}_j(\cdot|s,a)$.
    
    Assume to the contrary that for every $(s,a)\in\Sigma_{\hat{t}-1}^i$ we have $\mathcal{P}_i(\cdot|s,a)=\mathcal{P}_j(\cdot|s,a)$, i.e., the transition function is also equivalent for $t=\hat{t}-1$. Let $h_{\hat{t}}=(x,s_0,a_0,\ldots,s_{\hat{t}}, a_{\hat{t}})$ be the state-action history up to time $\hat{t}$, including the random variable $x$ that was used to choose a policy. We next consider the probability of observing a history under $(\mathcal{P}_i, \pi_r)$, 
    \begin{align*}
        P_{\mathcal{P}_i, \pi_r}(h_{\hat{t}}) =& P_{init}(s_0)P_x(x)\pi_{r}(a_0|x,s_0)P_{\mathcal{P}_i, \pi_r}(s_1|x,s_0, a_0)\pi_r(a_1|x,s_0,a_0,s_1)\cdots \\
        &\cdots P_{\mathcal{P}_i, \pi_r}(s_{\hat{t}}|x,s_0,a_0,\ldots,s_{\hat{t}-1}, a_{\hat{t}-1})\pi_r(a_{\hat{t}}|x,s_0,a_0,\ldots,s_{\hat{t}}) \\ =& P_{init}(s_0)P_x(x)\pi_{r}(a_0|x,s_0)\prod_{t=1}^{\hat{t}}{\mathcal{P}_i(s_t|s_{t-1}, a_{t-1})\pi_{r}(a_{t}|x,s_{0},a_{0},\ldots,s_{t})},
    \end{align*}
    where the last equality holds according to the Markov property, $P_{\mathcal{P}_i,\pi_r}(s_{t}|s_0,a_0,\ldots,s_{t-1}, a_{t-1}) = \mathcal{P}_i(s_t|s_{t-1}, a_{t-1})$. Since $\pi_r(a_t|x,s_0,a_0,\ldots,s_t)$ is the same replaying policy for all MDPs, and for every $t\leq\hat{t}-1$ and $(s,a)\in\Sigma_{t}^{i}$ we have $\mathcal{P}_i(\cdot|s,a)= \mathcal{P}_j(\cdot|s,a)$, then $P_{\mathcal{P}_i, \pi_r}(h_{\hat{t}})=P_{\mathcal{P}_j, \pi_r}(h_{\hat{t}})$. By marginalizing over $x,s_0,a_0,\ldots,s_{\hat{t}-1}, a_{\hat{t}-1}$ we obtain:
    \begin{align*}
        \sum_{x,s_0, a_0,\ldots,s_{\hat{t}-1}, a_{\hat{t}-1}}{P_{\mathcal{P}_i, \pi_r}(x,s_0,a_0,\ldots,s_{\hat{t}}, a_{\hat{t}})}&=\sum_{x,s_0, a_0,\ldots,s_{\hat{t}-1}, a_{\hat{t}-1}}{P_{\mathcal{P}_j, \pi_r}(x,s_0,a_0,\ldots,s_{\hat{t}}, a_{\hat{t}})} \\ P_{\mathcal{P}_i, \pi_r}(s_{\hat{t}}, a_{\hat{t}})&=P_{\mathcal{P}_j, \pi_r}(s_{\hat{t}}, a_{\hat{t}}),
    \end{align*}
    \vspace{-1em}
    which means that $\Sigma_{\hat{t}}^{i}=\Sigma_{\hat{t}}^{i,j}$, which contradicts the definition of $\hat{t}$.
\end{proof}
\newpage
\section{BOReL Pseudo-Code}\label{appendix:pseudo_code}
\begin{algorithm}
   \caption{BOReL}
   \label{alg:borel}
\begin{algorithmic}
   \STATE {\bfseries Input:} A set of MDPs $\{\mathcal{R}_i, \mathcal{P}_i\}_{i=1}^{N}\sim p(\mathcal{R}, \mathcal{P})$.
    \STATE \textcolor{gray}{\textit{Phase 1: Data Collection}}
    \FOR{$i=1,\ldots,N$}
        \STATE Train standard RL agent (e.g., DQN, SAC) to solve $\{\mathcal{R}_i, \mathcal{P}_i\}$
        \STATE Save the complete training history of the agent to buffer $\mathcal{D}_i$.
    \ENDFOR
    \STATE \textcolor{gray}{\textit{Phase 2 (Optional): Policy Replaying/Reward Relabelling}}
    \FOR{$i=1,\ldots,N$}
        \FOR{trajectory $\tau_k$ in $\mathcal{D}_i$}
        \IF{policy replaying}
            \STATE Uniformly draw a trained policy $\tilde{\pi}_k\sim\{\pi_{\beta}^{j}\}_{j=1}^{N}$.
            \STATE Collect trajectory $\tilde{\tau}_k$ by running $\tilde{\pi}_k$ on $\{\mathcal{R}_i, \mathcal{P}_i\}$.
            \STATE Replace $\tau_k$ by $\tilde{\tau}_k$.
        \ELSIF{reward relabelling}
            \STATE Uniformly draw $j\sim\{1,\ldots,N\}$.
            \STATE Replace every $r_{t+1}^{i,k}$ in $\tau_k$ with $\hat{r}_{t+1}^{i,k}=\mathcal{R}_j(s_t^{i,k}, a_t^{i,k})$.
        \ENDIF
        \ENDFOR
    \ENDFOR
    \STATE \textcolor{gray}{\textit{Phase 3: VAE Training and State Relabelling}}
    \STATE Train VAE using \eqref{eq:sum_elbo_obj} and data obtained at the previous step.
    \FOR{trajectory $\tau$ \textit{in} data}
        \FOR{$t=1,\ldots,H^{+}$}
        \STATE Pass $\tau_{:t}$ through encoder to obtain $b_t = \mu(\tau_{:t}),\Sigma(\tau_{:t})$ 
        \STATE Replace $s_t$ in trajectory with $s_t^{+} = (s_{t}, b_t)$.
    \ENDFOR
    \ENDFOR
    \STATE \textcolor{gray}{\textit{Phase 4: Offline Meta-RL Training}} 
    \STATE Train off-policy RL agent (e.g., DQN, SAC) using the offline data obtained from Phase 3.
\end{algorithmic}
\end{algorithm}
\section{VAE Training Objective}
For completeness, we follow \cite{zintgraf2020varibad} and outline the full training objective of the VAE. Consider the approximate posterior $q_{\phi}(m|h_{:t})$ conditioned on the history up to time $t$. In this case, the ELBO can be derived as follows:
\begin{align*}
    \log{P(s_0,r_1,s_1\ldots, s_H|a_0,\ldots,a_{H-1})} &= \log{\int P(s_0,r_1,s_1\ldots, s_H, m|a_0,\ldots,a_{H-1})dm} \\ &= \log{\int P(s_0,r_1,s_1\ldots, s_H, m|a_0,\ldots,a_{H-1})\frac{q_{\phi}(m|h_{:t})}{q_{\phi}(m|h_{:t})}dm} \\ &= \log{\mathbb{E}_{m\sim q_{\phi}(\cdot|h_{:t})}\left[\frac{P(s_0,r_1,s_1\ldots,s_H, m|a_0,\ldots,a_{H-1})}{q_{\phi}(m|h_{:t})}\right]} \\ &\geq \mathbb{E}_{m\sim q_{\phi}(\cdot|h_{:t})}\left[\log{p_{\theta}(s_0,r_1,s_1\ldots,s_H|m, a_0,\ldots,a_{H-1})}\right. \\ &\quad+ \log{p_{\theta}(m)} - \left.\log{q_{\phi}(m|h_{:t})}\right] \\ &= \mathbb{E}_{m\sim q_{\phi}(\cdot|h_{:t})}\left[ \log{p_\theta(s_0,r_1,s_1\ldots,s_H|m, a_0,\ldots,a_{H-1})}\right] \\ &\quad- D_{KL}(q_{\phi}(m|h_{:t}) || p_{\theta}(m)) \\ &=ELBO_t(\theta, \phi).
\end{align*}
The prior $p_{\theta}(m)$ is set to be the previous posterior $q_{\phi}(m|h_{:t-1})$, with initial prior chosen to be standard normal $p_{\theta}(m) = \mathcal{N}(0, I)$.  
The decoder $p_{\theta}(s_0,r_1,s_1\ldots,s_H|m, a_0,\ldots,a_{H-1})$ factorizes to reward and next state models $p_{\theta}(s'|s,a,m)$ and $p_{\theta}(r|s,a,m)$, according to:
\begin{align*}
    \log{p_{\theta}(s_0,r_1,s_1\ldots,s_H|m, a_0,\ldots,a_{H-1})} &= \log{p(s_0|m)} \\ &\quad+ \sum_{t=0}^{H-1}{\left[\log{p_{\theta}(s_{t+1}|s_t, a_t, m)} + \log{p_{\theta}(r_{t+1}|s_t, a_t, m)}\right]}.
\end{align*}

The overall training objective of the VAE is to maximize the sum of ELBO terms for different time steps,
\begin{equation} \label{eq:sum_elbo_obj}
    \max_{\theta, \phi}\sum_{t=0}^{H}{ELBO_t(\theta, \phi)}.
\end{equation}


\section{Environments Description}
\label{appendix:env_description}
In this section we describe the details of the domains we experimented with.
\paragraph{Gridworld:} A $5\times 5$ gridworld environment as in \cite{zintgraf2020varibad}. The task distribution is defined by the location of a goal, which is unobserved and can be anywhere but around the starting state at the bottom-left cell. For each task, the agent receives a reward of $-0.1$ on non-goal cells and $+1$ at the goal, i.e.,
\[
    r_t = \begin{cases}
    1, & s_t=g \\
    -0.1, & \text{else,}
    \end{cases}
\]
where $s_t$ is the current cell and $g$ is the goal cell. \\
Similarly to \cite{zintgraf2020varibad}, the horizon for this domain is set to $15$ and we aggregate $k=4$ consecutive episodes to form a trajectory of length $60$. 

\paragraph{Semi-circle:} A continuous 2D environment as in Figure \ref{fig:illustration}, where the agent must navigate to an unknown goal, randomly chosen on a semi-circle of radius $1$ \citep{rakelly2019efficient}. For each task, the agent receives a reward of $+1$ if it is within a small radius $r=0.2$ of the goal, and $0$ otherwise,
\[
    r_t = \begin{cases}
    1, & \Vert x_t-x_\text{goal}\Vert_2 \leq r \\
    0, & \text{else,}
    \end{cases} 
\]
where $x_t$ is the current 2D location. Action space is 2-dimensional and bounded: $\left[-0.1, 0.1\right]^2$. \\
We set the horizon to $60$ and aggregate $k=2$ consecutive episodes to form a trajectory of length $120$. 

\paragraph{MuJoCo:}
\begin{enumerate}
    \item \textbf{Half-Cheetah-Vel:} In this environment, a half-cheetah agent must run at a fixed target velocity. Following recent works in meta-RL \citep{finn2017model, rakelly2019efficient, zintgraf2020varibad}, we consider velocities drawn uniformly between $0.0$ and $3.0$. The reward in this environment is given by
    \[
        r_t = - |v_t - v_{\text{goal}}| - 0.05\cdot\Vert a_t\Vert_2^2
    \]
    where $v_t$ is the current velocity, and $a_t$ is the current action. The horizon is set to $200$ and we aggregate $k=2$ consecutive  episodes. 
    \item \textbf{Ant-Semi-circle:} In this environment, an ant needs to navigate to an unknown goal, randomly chosen on a semi-circle, similarly to the Semi-circle task above. 
    
    When collecting data for this domain, we found that the standard SAC algorithm \citep{haarnoja2018soft} was not able to solve the task effectively due to the sparse reward (which is described later), and did not produce trajectories that reached the goal. We thus modified the reward \textbf{only during data collection} to be dense, and inversely proportional to the distance from the goal,
    \[
        r_t^{\text{dense}} = -\Vert x_t - x_{\text{goal}}\Vert_1 - 0.1\cdot\Vert a_t\Vert_2^2
    \]
    where $x_t$ is the current $2$D location and $a_t$ is the current action.
    After collecting the data trajectories, we replaced all the dense rewards in the data with the sparse rewards that are given by
    \[
        r_t^{\text{sparse}} = - 0.1\cdot\Vert a_t\Vert_2^2 + \begin{cases}
        $1$, & \Vert x_t - x_{\text{goal}}\Vert_2 \leq 0.2 \\ $0$, & \text{else.}
        \end{cases}
    \]
    We note that \cite{rakelly2019efficient} use a similar approach to cope with sparse rewards in the online setting. \\ The horizon is set to $200$ and we aggregate $k=2$ consecutive episodes. 

    \item \textbf{Reacher-Image:}  In this environment, a two-link planar robot needs to reach an unknown goal, randomly chosen on a quarter circle. The robot receives dense reward which is given by
    \[
        r_t = - \Vert x_t - x_{\text{goal}}\Vert_2
    \]
    where $x_t$ is the location of the robot's end effector. The agent observes single-channel images of size $64\times 64$ of the environment (see Figure~\ref{fig:trained_agents_vis}b). The horizon is set to $100$ and we aggregate $k=2$ consecutive  episodes. 
\end{enumerate}

\paragraph{Wind:} A continuous 2D domain with varying transitions, where the agent must navigate to a fixed (unknown) goal within a distance of $D=1$ from its initial state (the goal location is the same for all tasks). Similarly to Semi-circle, the agent receives a reward of $+1$ if it is within a radius $r=0.2$ of the goal, and $0$ otherwise, 
\[
    r_t = \begin{cases}
    1, & \Vert s_t-s_\text{goal}\Vert_2 \leq r \\
    0, & \text{else.}
    \end{cases} 
\]
For each task in this domain, the agent is experiencing a different `wind', which results in a shift in the transitions, such that when taking action $a_t\in\left[-0.1, 0.1\right]^2$ from state $s_t$ in MDP $\mathcal{M}$, the agent transitions to a new state $s_{t+1}$, which is given by
\[
    s_{t+1} = s_t + a_t + w_{\mathcal{M}},
\]
where $w_{\mathcal{M}}$ is a task-specific wind, which is randomly drawn for each task from a uniform distribution over $\left[-0.05, 0.05\right]^2$. To navigate correctly to the goal and stay there, the agent must take actions that cancel the wind effect. \\
We set the horizon to $25$ and evaluate the performance in terms of average return within the \textbf{first} episode of interaction on test tasks ($k=1$). 

\paragraph{Escape-Room:} A continuous 2D domain where the agent must navigate outside a circular room of radius $R=1$ through an opening, whose location is unknown. For all tasks, the central angle of the opening is $\pi/8$. The tasks differ by the location of the opening -- the center point of the opening is sampled uniformly from $[0, \pi]$. The reward function is sparse, task-independent, and given by
\[
    r_t = \begin{cases}
    1, & \Vert s_t\Vert_2 > R \\
    0, & \text{else.}
    \end{cases} 
\]
The transition function, however, is task-dependent and given by

\[
    s_{t+1} = \begin{cases}
    \frac{s_{t}+a_{t}}{\lVert s_{t}+a_{t}\rVert_2}, & \text{if \textit{intersection occurs}} \\
    s_{t}+a_{t}, & \text{else,}
    \end{cases} 
\]
where \textit{intersection occurs} means that the line that connects $s_t$ and $s_t+a_t$ and the wall of the circular room intersects.
To solve a task, the agent must search for the opening by moving along the wall until he finds it. \\
We set the horizon to $60$ and aggregate $k=2$ consecutive episodes to a form a trajectory of length $120$.

\section{Experimental Details}
In this section we outline our training process and hyperparameters. \\ \\
For the discrete Gridworld domain we used DQN \citep{mnih2015human} with soft target network updates, as proposed by \cite{lillicrap2015continuous}, which has shown to improve the stability of learning. For the rest of the continuous domains, we used SAC \citep{haarnoja2018soft} with the architectures of the actor and critic chosen similarly, and with a fixed entropy coefficient. For both DQN and SAC, we set the soft target update parameter to $0.005$. \\ \\ 
In our experiments we average performance over $3$ random seeds and present the mean and standard deviation. \\
Our offline training procedure is comprised of $3$ separate training steps. First is the training of the data collection RL agents. Each agent is trained on a different task from the task distribution. 

For all domains but Reacher-Image, we used a similar architecture of $2$ fully-connected (FC) hidden layers of size that depends on the domain with ReLU activations, and set the batch size to $256$. 

For Reacher-Image, we used data augmentation techniques as suggested by \citet{laskin2020reinforcement}. Specifically, we used random translations and cropping. Then, we pass the observation through a convolutional neural network (CNN) with $4$ hidden layers followed by $2$ FC hidden layers. 

The rest of the hyperparameters used for training the data collection RL agents are summarized in the following table:
\begin{center}
\begin{tabular}{ |l|c|c|c|c|c|c|c| } 
\hline
& \textbf{Gridworld}  & \textbf{Semi-circle} & \textbf{Cheetah \& Ant} & \textbf{Reacher} & \textbf{Wind} & \textbf{Escape-Room} \\
\specialrule{.1em}{.05em}{.05em}
\textbf{Num. train tasks} & 21 & 80 & 100 \& 80 & 50 & 40 & 60 \\
Hidden layers size & 16 & 32 & 128 & 1024 & 64 & 128 \\
Num. iterations & $200$ & $300$ & $1000$ & $50$ & $300$ & $50$ \\
RL updates per iter. & $500$ & $500$ & $2000$ & $500$ & $500$ & $500$ \\
\multirow{3}{*}{\vtop{\hbox{\strut Exploration/} \hbox{\strut entropy coeff.}}} & \vtop{\hbox{\strut $\epsilon$-greedy, annealing}\hbox{\strut from $1$ to $0.1$} \hbox{\strut over $100$ iterations}} & \multirow{3}{*}{$0.01$} & \multirow{3}{*}{$0.2$} & \multirow{3}{*}{$0.05$} & \multirow{3}{*}{$0.01$} & \multirow{3}{*}{$0.01$} \\   
Collected ep. per iter. & $5$ & $2$ & $2$ & $1$ & $2$ & $2$ \\
Learning rate/s & $3\cdot 10^{-4}$ & $3\cdot 10^{-4}$ & $3\cdot 10^{-4}$ & $1\cdot 10^{-3}$ & $3\cdot 10^{-4}$ & $3\cdot 10^{-4}$ \\
Discount factor ($\gamma$) & $0.99$ & $0.9$ & $0.99$ & $0.99$ & $0.9$ & $0.9$ \\
\hline
\end{tabular}
\end{center}
The second training step is the VAE training after optionally applying reward relabelling/policy replaying to the collected data.

The VAE consists of a recurrent encoder, which at time step $t$ takes as input the tuple $(a_t, r_{t+1}, s_{t+1})$. The state and reward are passed each through a different fully-connected (FC) layer (preceded by a CNN feature-extractor in Reacher-Image). The state FC layer is of size $32$ and the reward FC layer is of size $8$ for the Gridworld and $16$ for the rest of the domains, all with ReLU activations. For all environments but Gridworld, we also pass the action through a FC layer of size $16$ with ReLU. Then, the state and reward layers' outputs are concatenated along with the action (or with the output of the action layer) and passed to a GRU of size $64/128$ (Gridworld/other domains). The GRU outputs the Gaussian parameters $\mu(h_{:t}), \Sigma(h_{:t})$ of the latent vector $m$, whose dimensionality is $5$ in all our experiments. 
 
For all reward-varying domains (all but Wind/Escape-Room), we only train reward-decoder (Similarly to \citet{zintgraf2020varibad}). For Wind and Escape-room we also train transition decoder. In all domains, the decoder network/s are comprised of $2$ FC layers, each of size $32$.

The VAE is trained to optimize Equation~$\eqref{eq:sum_elbo_obj}$, but similarly to \cite{zintgraf2020varibad}, we weight the KL term in each of the ELBO terms with some parameter $\beta$, which is not necessarily $1$. 
In our experiments we used $\beta=0.05$. 

After the VAE is trained, we apply state relabelling to the data collected by the RL agents, to create a large offline dataset that effectively comes from the BAMDP. Then, we train an off-policy RL algorithm, which is our meta-RL agent, using the offline data. 

For the offline meta-RL agents training, we used similar hyperparameters to those used for the data collection RL agents training. For some of the domains, we enlarge the size of the hidden layers.

\section{Learned Belief and Policy Visualization}\label{appendix:learned_belief}
In this section we visualize the learned belief states, in order to get more insight into the decision making process of the agent during interaction. We also visualize trajectories of trained agents in different domains.

\begin{figure}[h]
    \centering
    \begin{subfigure}{0.85\textwidth}
        \centering
        \includegraphics[width=\linewidth]{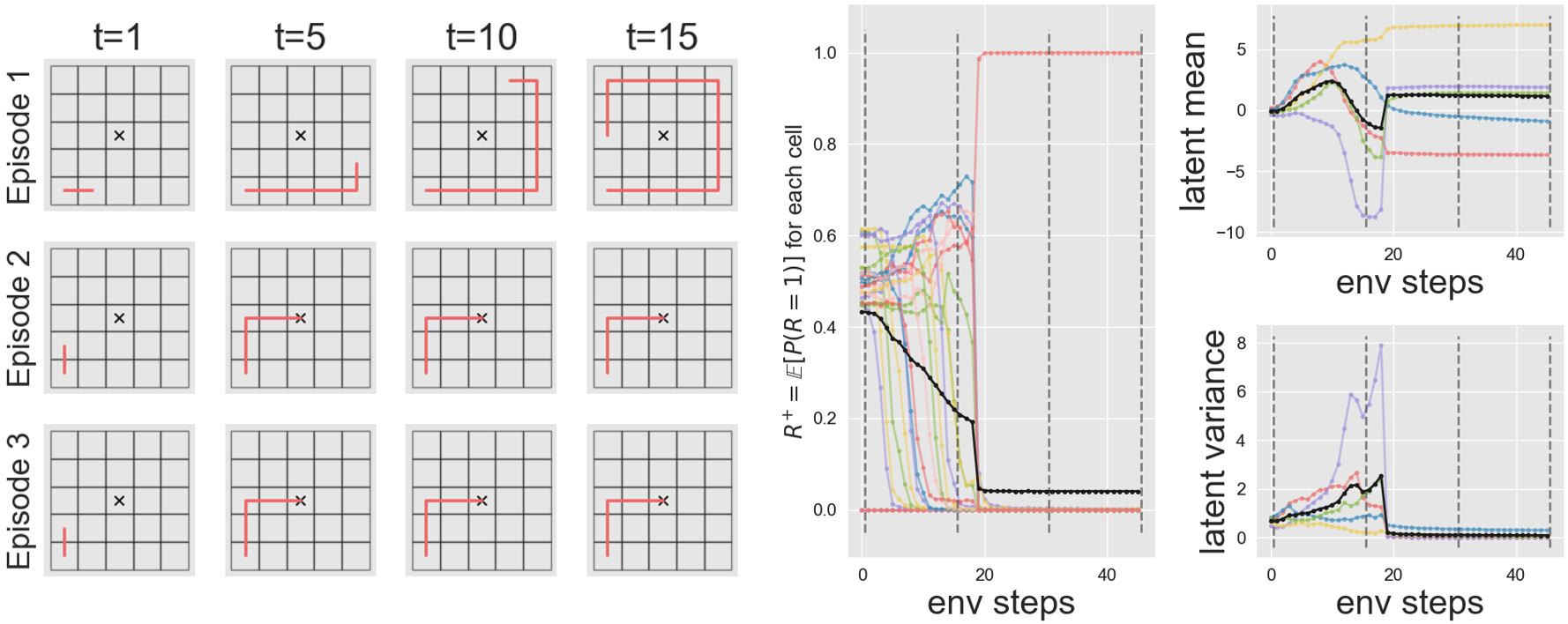} 
        \label{fig:belief_gridworld}
        \vspace{-2em}
        \subcaption{Gridworld}
    \end{subfigure}\vspace{1em}
    \begin{subfigure}{0.48\textwidth}
        \centering
        \includegraphics[width=\linewidth]{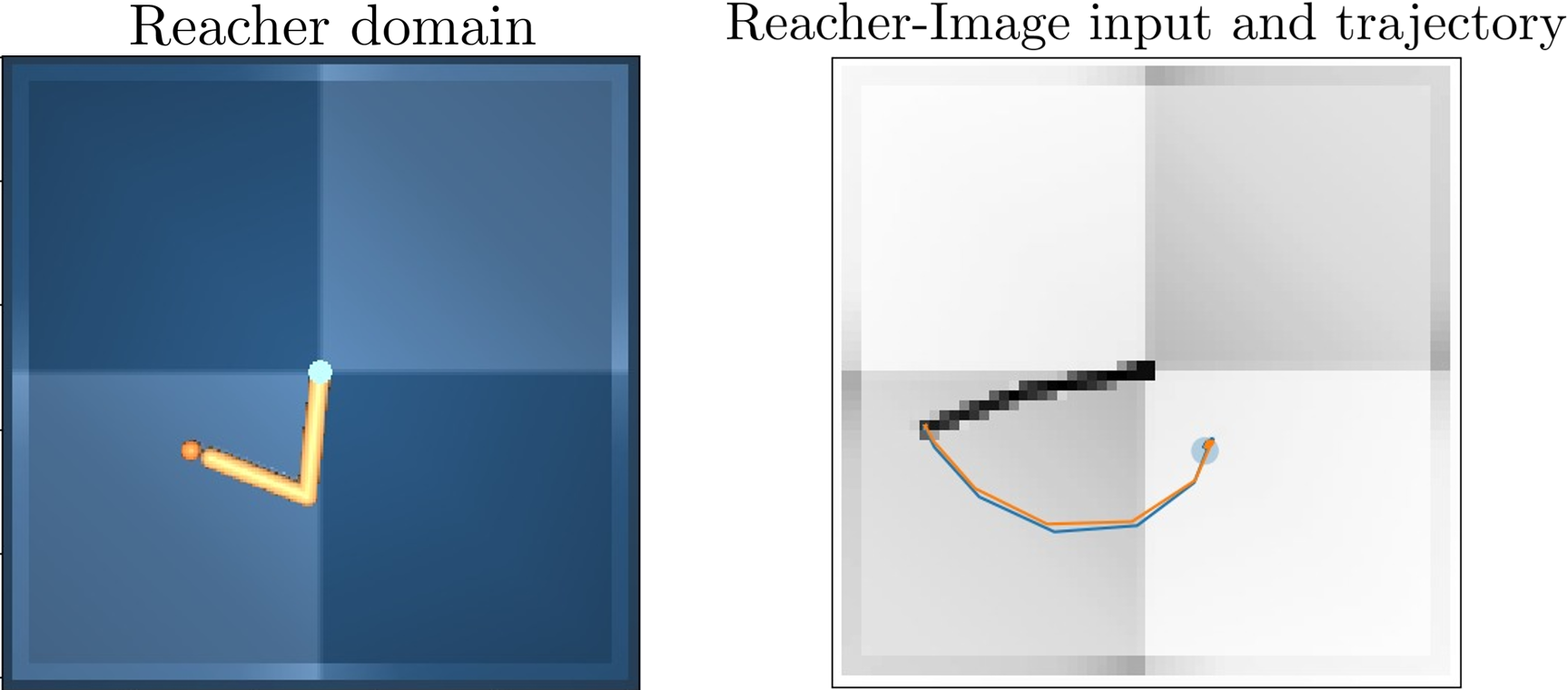} 
        \label{fig:reacher_image}
        \vspace{-1em}
        \subcaption{Reacher-Image}
    \end{subfigure}
    \begin{subfigure}{0.48\textwidth}
        \centering
        \includegraphics[width=\linewidth]{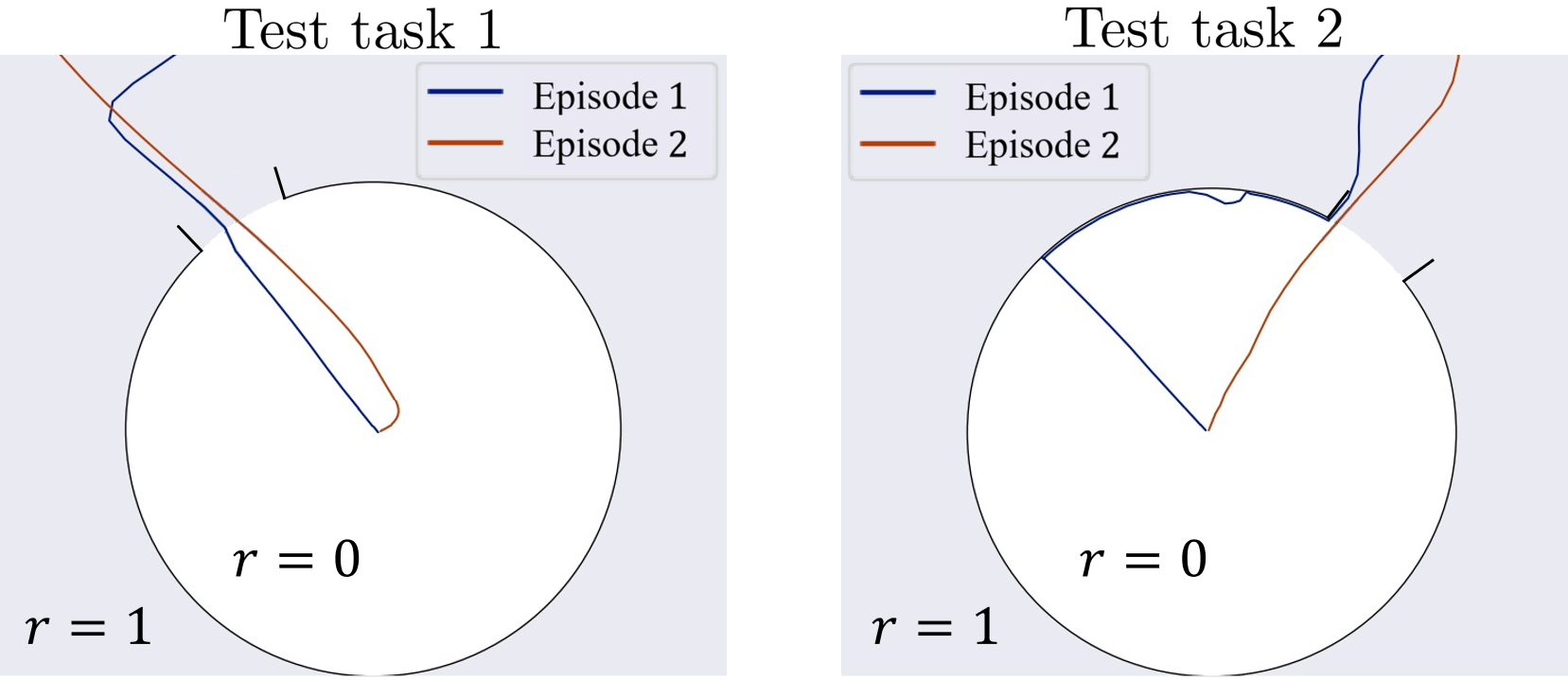} 
        \label{fig:escape_room}
        \vspace{-1em}
        \subcaption{Escape-Room}
    \end{subfigure}\vspace{1em}
    \begin{subfigure}{0.8\textwidth}
        \centering
        \includegraphics[width=\linewidth]{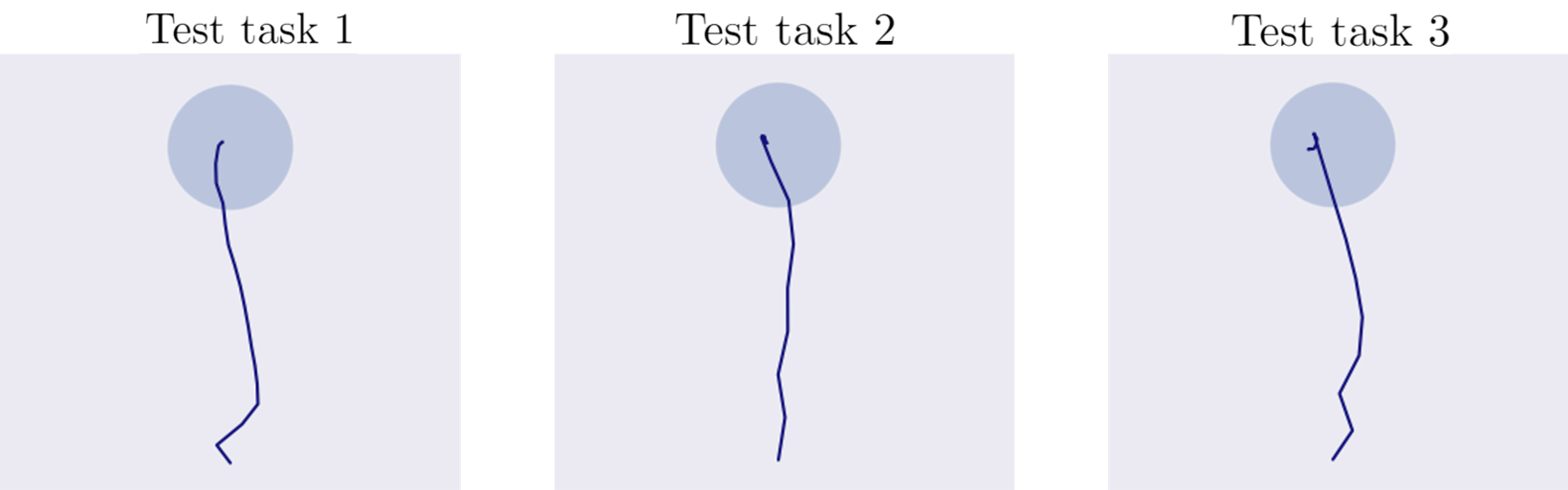} 
        \label{fig:wind}
        \vspace{-1em}
        \subcaption{Wind}
    \end{subfigure}
    \caption{Interaction of trained agents with evaluated domains. In (a) we show interaction with Gridworld, including belief update throughout interaction (for more details, see \citet{zintgraf2020varibad}). In (b), (c) and (d) we show typical behavior of trained agents interacting with Reacher-Image, Escape-Room and Wind, respectively.}
    \label{fig:trained_agents_vis}
\end{figure}

In Figure~\ref{fig:trained_agents_vis}a, we visualize the interaction of a trained agent with the Gridworld environment, exactly as visualized in Figure 3 at \cite{zintgraf2020varibad}. The agent reduces its uncertainty by effectively searching the goal. After the goal is found, the agent stops and in subsequent episodes it directly moves toward it.  

The Reacher domain is visualized in Figure~\ref{fig:trained_agents_vis}b. In the left side, an RGB image of the domain is presented. In the right side, we present the input image to the agent (which consists of a single-channel and has lower resolution) along with successful trajectories that reaches a goal from the test set.

In Figure~\ref{fig:trained_agents_vis}c we show the typical behavior of a trained agent in Escape-Room domain. Note that in 'Test task 1' the agent finds the opening without colliding with the walls of the room and in the second episode the agent follows a similar trajectory that leads to reward. On the other hand, in 'Test task 2' the agent collides with the wall in the first episode, and then it effectively searches for the opening. After he finds it, in the second episode he directly escapes the room.

In Figure~\ref{fig:trained_agents_vis}d we visualize trajectories of a trained agent on different test tasks in Wind domain. As can be seen, after several steps in the environment, our agent learns to adapt to the varying wind, and travels to the goal in a straight line. PEARL, on the other hand, only adapts after the first episode, and therefore obtains worse results (see Figure~\ref{fig:offline_res_trnasitions}). We believe it is possible to improve PEARL to update its posterior after every step, and in this case the improved PEARL will obtain similar performance as our method in Wind. However, this will not work in the sparse domains described in the main text, where the Bayes adaptive exploration has an inherent advantage over Thompson sampling. We emphasize that in Wind, MDP ambiguity is not a concern, since the data from all agents is largely centered on the line between the agent's initial position and the goal. Thus, the effect of the wind on these states can uniquely be identified in each task. 

\begin{figure}
    \centering
    \includegraphics[width=0.8\linewidth]{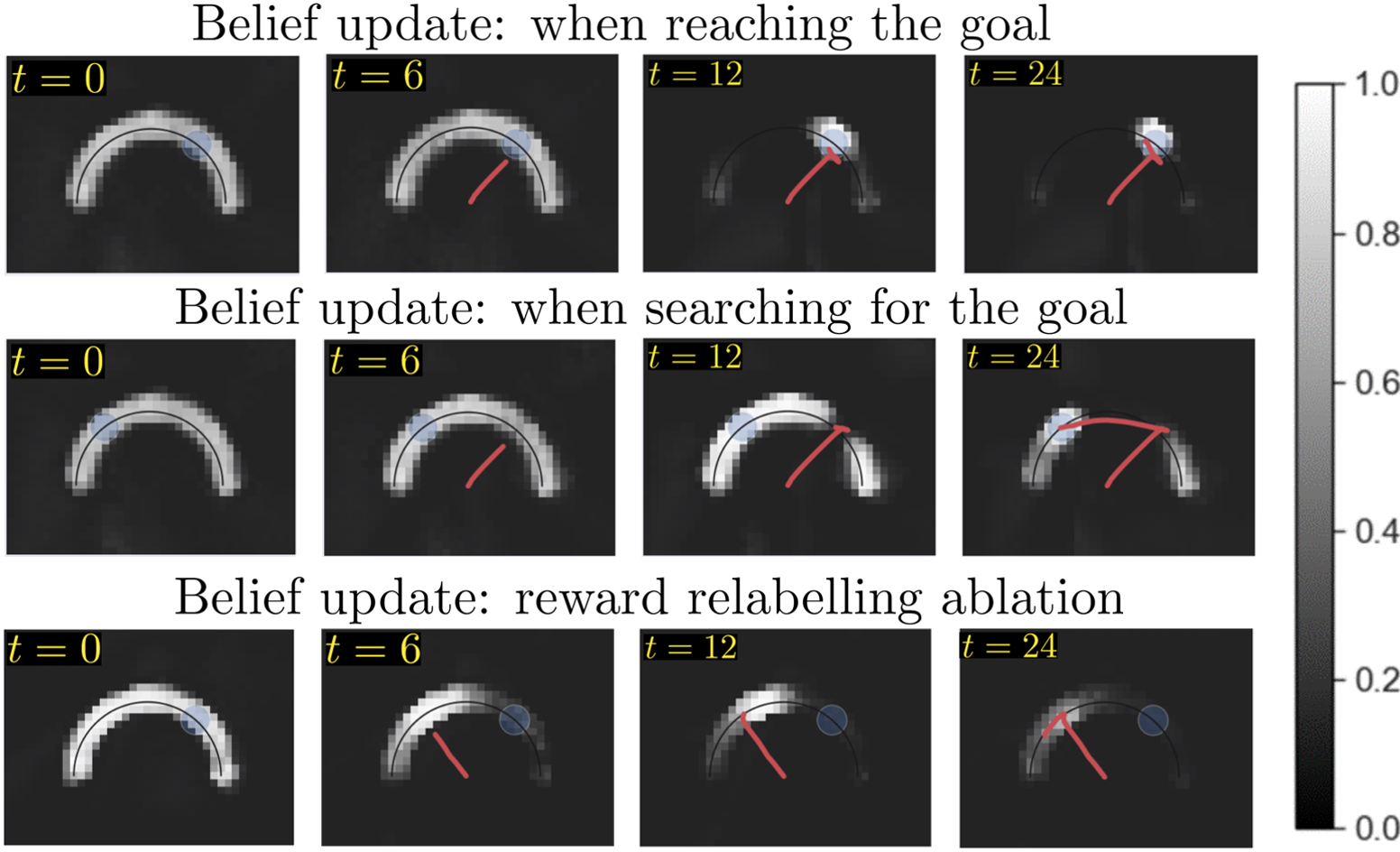}
    \caption{Semi-circle belief visualization. The plots show the reward belief over the 2-dimensional state space (obtained from the VAE) at different stages of interacting with the system. The red line marks the agent trajectory, and the light blue circle marks the true reward location. \textbf{Top:} Once the agent finds the true goal, it reduces the belief over other possible goals from the task distribution. \textbf{Middle:} As long as the agent doesn't find the goal, it explores efficiently, reducing the uncertainty until the goal is found. \textbf{Bottom:} Without reward relabelling, the agent doesn't learn to differentiate between different MDPs, and therefore fails to identify the goal.}
    \label{fig:belief_halfcircle}
    \vspace{-1em}
\end{figure}

In Figure \ref{fig:belief_halfcircle}, we plot the reward belief (obtained from the VAE decoder) at different steps during the agent's interaction in the Semi-circle domain. Note how the belief starts as uniform over the semi-circle, and narrows in on the target as more evidence is collected.
Also note that without reward relabelling, the agent fails to find the goal. In this instance of the MDP ambiguity problem, the training data for the meta-RL agent consists of trajectories that mostly reach the goal, and as a result, the agent believes that the reward is located at the first point it reaches on the semi-circle. 

\newpage
\section{Data Quality Ablation}
\label{appendix:data_quality}
\begin{figure}[ht]
    \begin{subfigure}{0.33\textwidth}
        \centering
        \includegraphics[width=\linewidth]{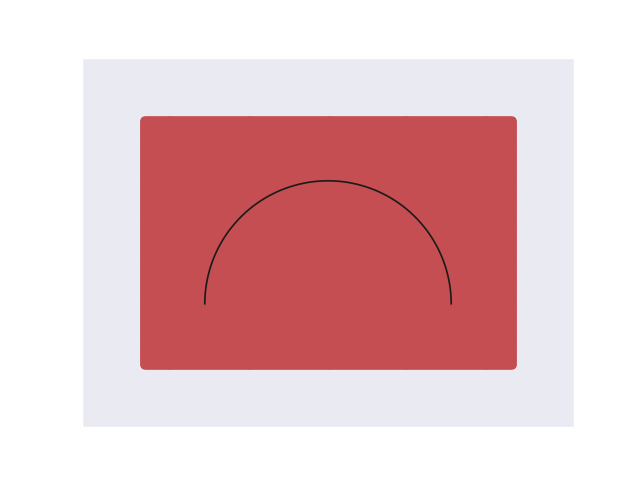} 
        \label{fig:data_uni}
        \vspace{-2em}
        \caption{Uniform}
    \end{subfigure}
    \begin{subfigure}{0.33\textwidth}
        \centering
        \includegraphics[width=\linewidth]{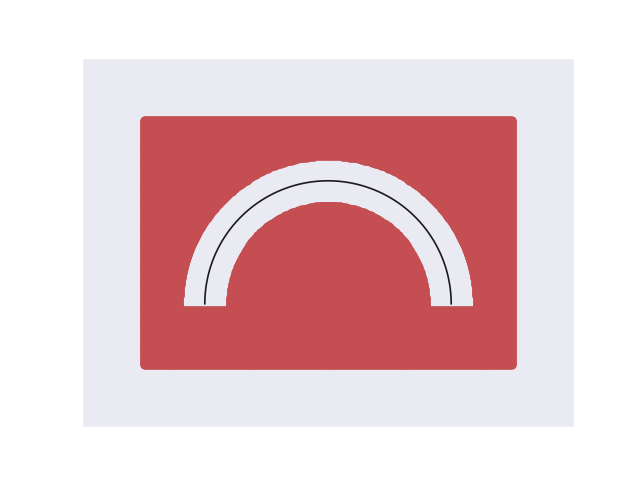} 
        \label{fig:data_ex_sc}
        \vspace{-2em}
        \caption{Excluding s.c.}
    \end{subfigure}
    \begin{subfigure}{0.33\textwidth}
        \centering
        \includegraphics[width=\linewidth]{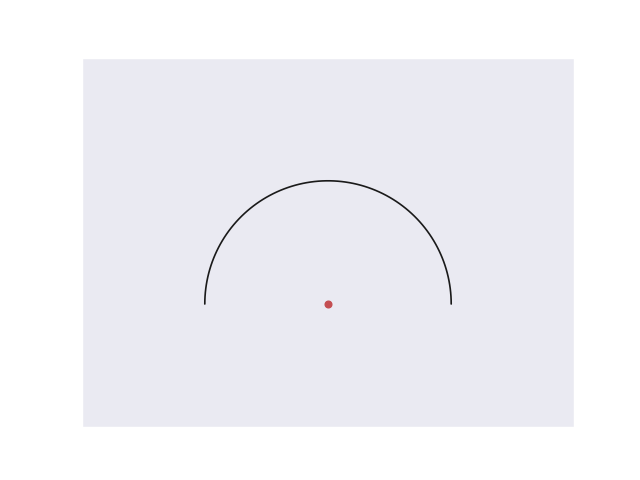} 
        \label{fig:data_sp}
        \vspace{-2em}
        \caption{Fixed}
    \end{subfigure}
    \caption{Initial state distributions. Red locations indicate non-zero sampling probability.}
    \label{fig:initial_state_dists}
\end{figure}

In our data quality ablative study, we consider the Ant-Semi-circle domain for which we modify the initial state distribution during the data collection phase. The initial state distributions we consider are visualized in Figure~\ref{fig:initial_state_dists}: Uniform distribution, uniform excluding states on the semi-circle, and fixed initial position.
\vspace{-1em}

\begin{figure}[ht]
    \begin{subfigure}{0.3\textwidth}
        \centering
        \includegraphics[width=\linewidth]{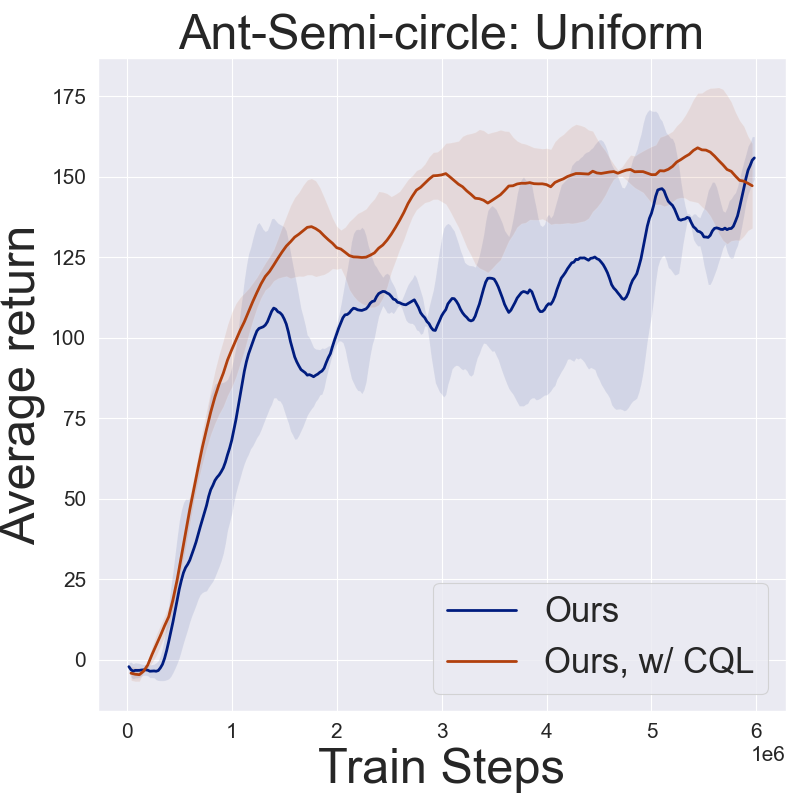} 
        \label{fig:data_conv_uniform}
    \end{subfigure}\hfill
    \begin{subfigure}{0.3\textwidth}
        \centering
        \includegraphics[width=\linewidth]{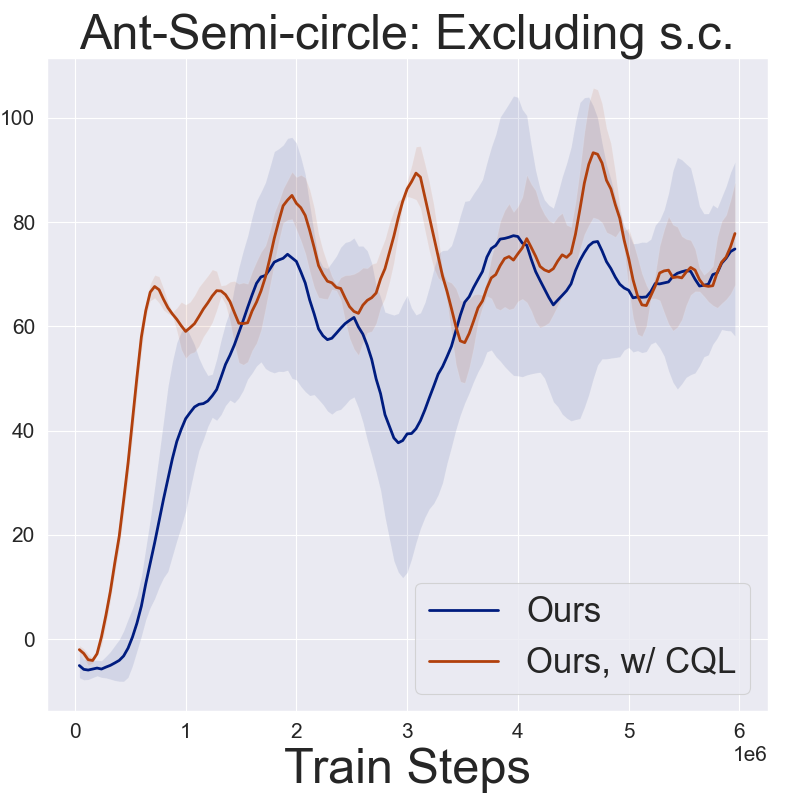} 
        \label{fig:data_conv_ex_sc}
    \end{subfigure}\hfill
    \begin{subfigure}{0.3\textwidth}
        \centering
        \includegraphics[width=\linewidth]{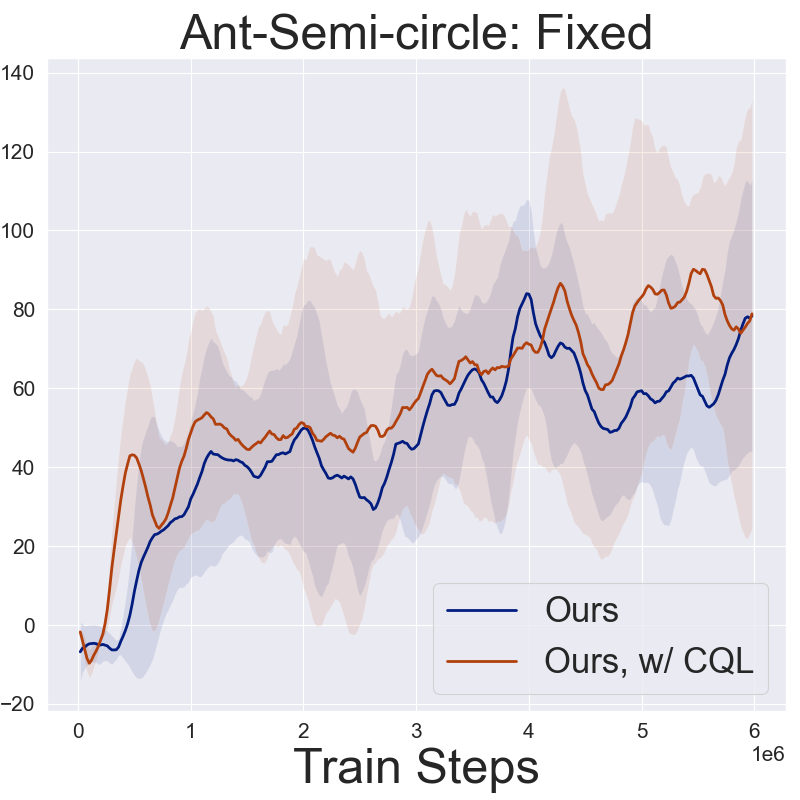} 
        \label{fig:data_conv_sp}
    \end{subfigure}
    \vspace{-1em}
    \caption{Learning curves for the results presented in Table~\ref{wrap-tab:1}. In blue is our method and in red is our method with critic network trained according to the CQL objective \citep{kumar2020conservative}. \textbf{Left:} Uniform initial state distribution. \textbf{Middle:} Uniform distribution, excluding states over the semi-circle. \textbf{Right:} Initial state is fixed.}
    \label{fig:initial_state_learn_curves}
\end{figure}

Figure~\ref{fig:initial_state_learn_curves} shows the learning curves for the results presented in Table \ref{wrap-tab:1}. For completeness, we add the learning curve for the uniform distribution which is also presented in Figure \ref{fig:offline_res}. 

\begin{figure}[ht]
    \centering
        \includegraphics[width=0.92\linewidth]{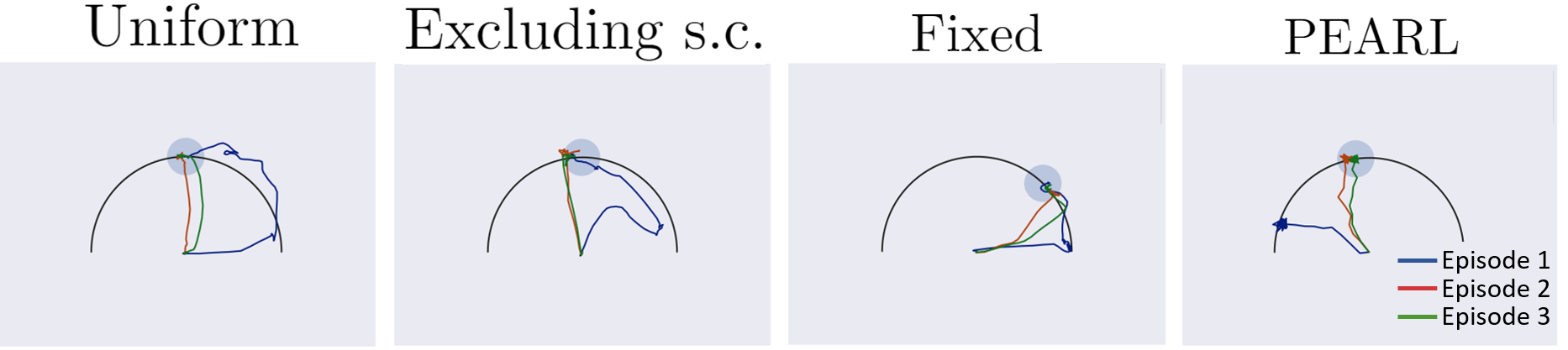} 
    \caption{Ant-Semi-circle: trajectories of trained agents for different offline datasets and for PEARL.
    }
    \label{fig:visualize_trajs_initial_dist}
\end{figure}

We also visualize trajectories of trained agents for the $3$ different cases as well as for PEARL \citep{rakelly2019efficient}, in Figure~\ref{fig:visualize_trajs_initial_dist}. Note that even for the fixed-distribution dataset, our agent learns to search for the goal.

\section{Online Setting Performance}
\begin{figure}[h]
    \centering
    \includegraphics[width=0.85\textwidth]{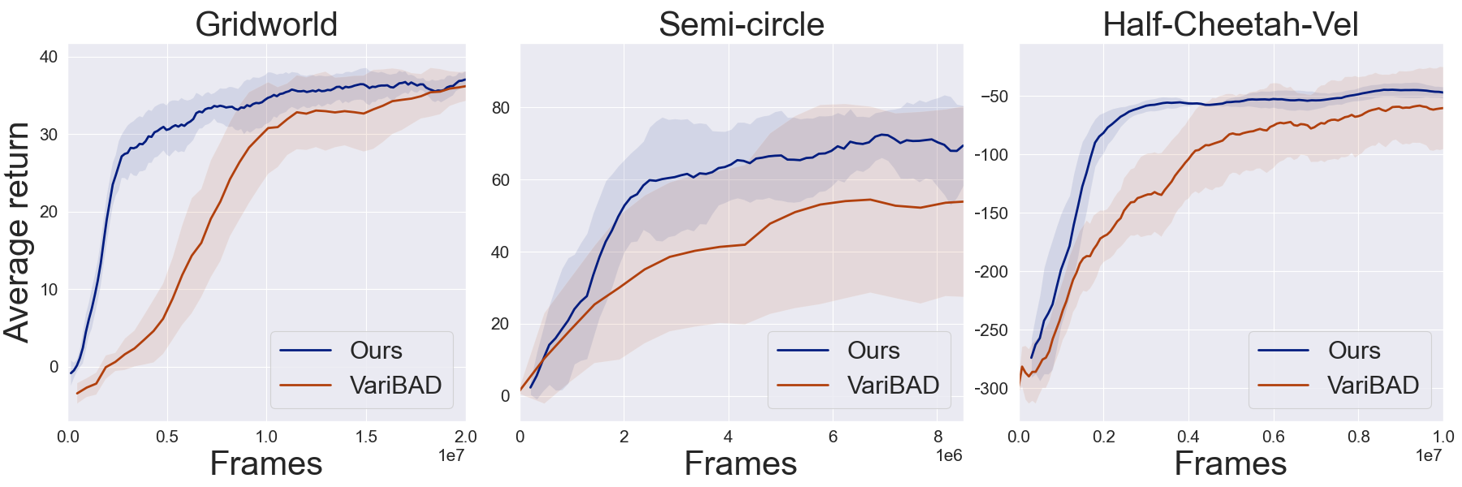}
    \caption{Online performance comparison. The off-policy optimization significantly improved VariBAD performance.}
    \label{fig:online_res}
\end{figure}

Our method can also be applied to the online setting, in which online data collection is allowed. In this case, it is simply a modification of VariBAD, where the policy gradient optimization is replaced with an off-policy RL algorithm. Since MDP ambiguity does not concern online meta-RL, we did not use reward relabelling in this setting.
As shown in Figure~\ref{fig:online_res}, by exploiting the efficiency of off-policy RL, our method significantly improves sample-efficiency, without sacrificing final performance. 

When comparing Figure \ref{fig:online_res} and Figure \ref{fig:offline_res}, the reader may notice that the online algorithm's final performance outperforms the final performance in the offline setting. We emphasize that this phenomenon largely depends on the quality of the offline data, and not on the algorithm itself.

The hyperparameters used in the online setting are as follows:

\begin{center}
\begin{tabular}{ |l|c|c|c|c| } 
\hline
& \textbf{Gridworld (DQN)}  & \textbf{Semi-circle (SAC)} & \textbf{Cheetah-Vel (SAC)} \\
\specialrule{.1em}{.05em}{.05em}
\textbf{RL parameters} & & & \\
\multirow{2}{*}{Architecture/s} & \vtop{\hbox{\strut $2$ FC layers }\hbox{\strut of size $100$.}} & \vtop{\hbox{\strut $2$ FC layers }\hbox{\strut of size $128$.}} & \vtop{\hbox{\strut $3$ FC layers }\hbox{\strut of size $128$.}} \\
Num. updates per iter. & $250$ & $1000$ & $2000$ \\
\multirow{3}{*}{Exploration/entropy coeff.} 
& \vtop{\hbox{\strut $\epsilon$-greedy, linear }\hbox{\strut annealing from $1$ to $0.1$ } \hbox{\strut over $1000$ iterations.}} & \multirow{3}{*}{$0.01$} & \multirow{3}{*}{$0.2$} \\   
Collected episodes per iter. & $25$ & $25$ & $25$ \\
Learning rate/s & $7\cdot 10^{-5}$ & $7\cdot 10^{-5}$ & $3\cdot 10^{-4}$ \\
Discount factor ($\gamma$) & $0.99$ & $0.9$ & $0.99$ \\
\specialrule{.1em}{.05em}{.05em}
\textbf{VAE parameters} & & & \\
\multirow{3}{*}{Encoder architecture} & \vtop{\hbox{\strut state/reward FC layer} \hbox{\strut of size $32/8$.} \hbox{\strut GRU of size $64$.}} &\vtop{\hbox{\strut state/reward FC layer} \hbox{\strut of size $32/8$.} \hbox{\strut GRU of size $128$.}} &\vtop{\hbox{\strut state/action/reward FC} \hbox{\strut layer of size $32/16/16$.} \hbox{\strut GRU of size $128$.}} \\
\multirow{2}{*}{Reward decoder architecture} & \vtop{\hbox{\strut $2$ FC layers } \hbox{\strut of size $32$.}} & \vtop{\hbox{\strut $2$ FC layers of} \hbox{\strut sizes $64$ and $32$.}} & \vtop{\hbox{\strut $2$ FC layers of} \hbox{\strut sizes $64$ and $32$.}} \\
Num. updates per iter. & 20 & 25 & 20 \\
Learning rate & $3\cdot 10^{-4}$ & $10^{-3}$ & $3\cdot 10^{-4}$ \\
Weight of KL term $(\beta)$ & $1.0$ & $0.1$ & $1.0$ \\
\hline
\end{tabular}
\end{center}

\section{Additional Results}
\subsection{Performance vs. Adaptation Episodes}
\begin{figure}[ht]
    \centering
    \includegraphics[width=0.55\textwidth]{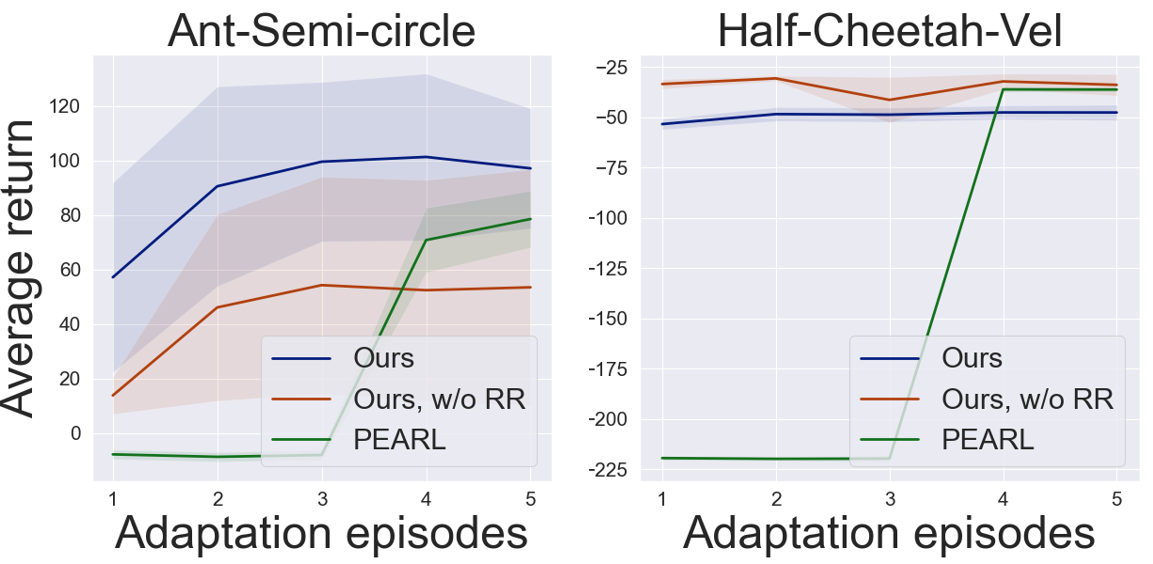}
    \caption{Adaptation performance. Our method outperforms PEARL, collecting high rewards within the first adaptation episodes.}
    \label{fig:performance_vs_episodes}
\end{figure}

In this part, we present the average reward per-episode as a function of the number of adaptation episodes at the environment. 
Figure~\ref{fig:performance_vs_episodes} shows the performance for the Ant-Semi-circle and Half-Cheetah-Vel domains. Note that within the first few episodes, PEARL does not collect high rewards due to the Thompson sampling-based nature of the algorithm. Our method, on the other hand, efficiently explores new tasks and is able to collect rewards within the first episodes of interaction.

\subsection{PEARL Learning Curves}
\begin{figure}[ht]
    \begin{subfigure}{0.3\textwidth}
        \centering
        \includegraphics[width=\linewidth]{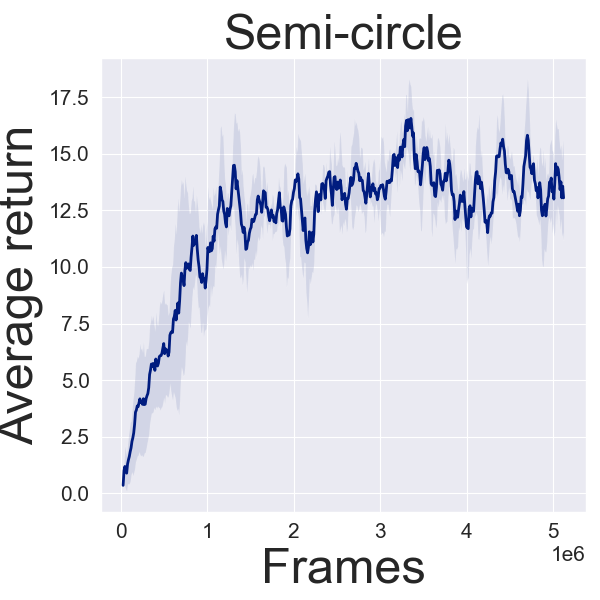} 
        \label{fig:pearl_semi_circle}
    \end{subfigure}\hfill
    \begin{subfigure}{0.3\textwidth}
        \centering
        \includegraphics[width=\linewidth]{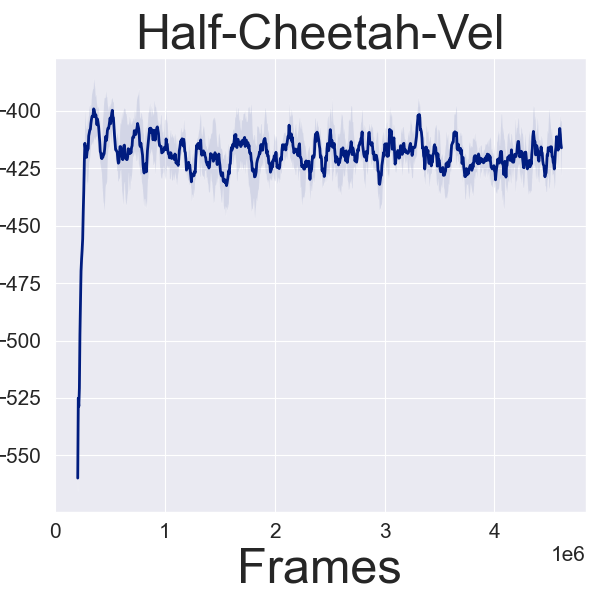} 
        \label{fig:pearl_half_cheetah}
    \end{subfigure}\hfill
    \begin{subfigure}{0.3\textwidth}
        \centering
        \includegraphics[width=\linewidth]{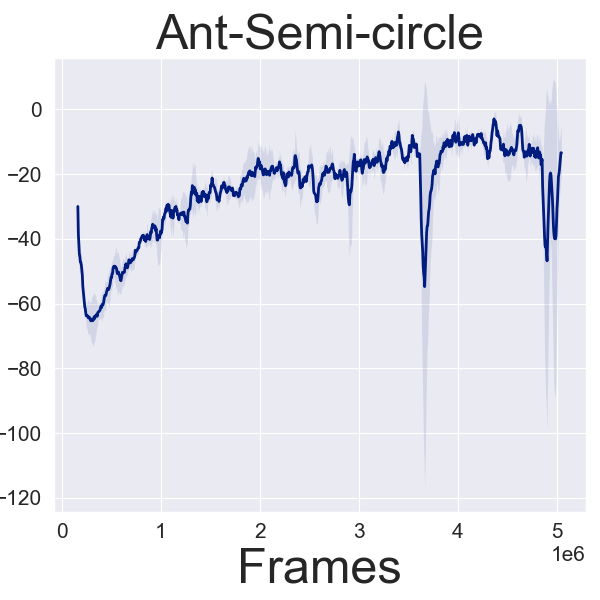} 
        \label{fig:pearl_ant}
    \end{subfigure}
        \begin{subfigure}{0.3\textwidth}
        \centering
        \includegraphics[width=\linewidth]{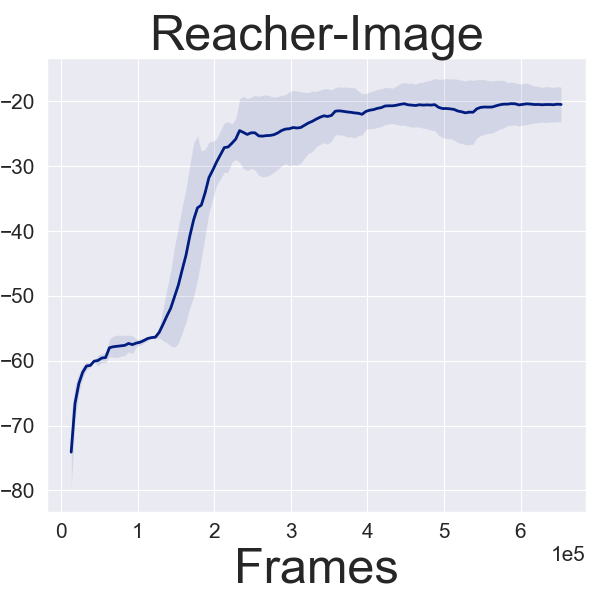} 
        \label{fig:pearl_reacher}
    \end{subfigure}\hfill
    \begin{subfigure}{0.3\textwidth}
        \centering
        \includegraphics[width=\linewidth]{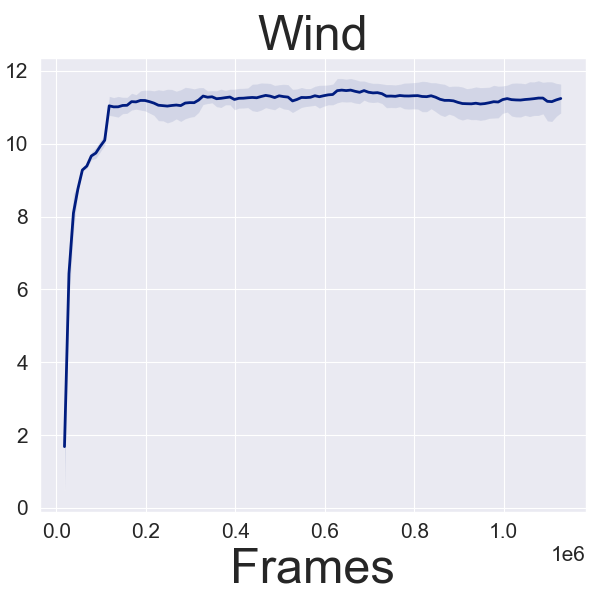} 
        \label{fig:pearl_wind}
    \end{subfigure}\hfill
    \begin{subfigure}{0.3\textwidth}
        \centering
        \includegraphics[width=\linewidth]{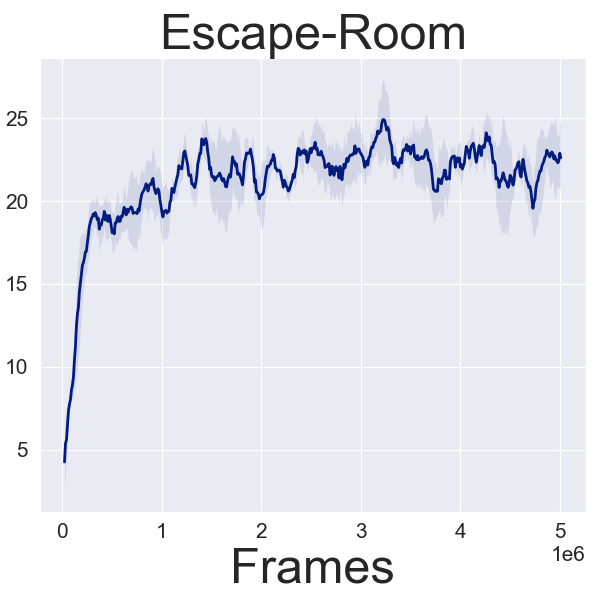} 
        \label{fig:pearl_escape_room}
    \end{subfigure}
    \vspace{-1em}
    \caption{Learning curves for online PEARL training.} 
    \label{fig:pearl_train_curves}
\end{figure}

We present the training curves of PEARL in Figure~\ref{fig:pearl_train_curves}. Note that since PEARL is an online algorithm, the $x$-axis represents the number of environment interactions.

\end{document}